\documentclass{article}

\usepackage[preprint]{neurips_2025}

\usepackage[utf8]{inputenc} 
\usepackage[T1]{fontenc}    
\usepackage{hyperref}       
\usepackage{url}            
\usepackage{booktabs}       
\usepackage{amsfonts}       
\usepackage{graphicx}       
\usepackage{nicefrac}       
\usepackage{microtype}      
\usepackage{xcolor}         
\usepackage{xspace}
\usepackage{adjustbox}
\usepackage{ulem}
\usepackage{listings}
\usepackage{array}
\usepackage{longtable}
\usepackage{ragged2e}
\usepackage{seqsplit}
\usepackage{adjustbox}
\usepackage{multirow}
\usepackage{tabularx}
\usepackage{amssymb}
\usepackage{amsmath}
\usepackage{pifont}
\usepackage{caption} 
\usepackage{subcaption}
\usepackage{wrapfig}
\usepackage{tikz}
\usepackage{comment}
\usepackage{pifont}

\usepackage{pgf-pie}
\newcommand*\emptycirc[1][1ex]{\tikz\draw (0,0) circle (#1);} 
\newcommand*\halfcirc[1][1ex]{%
	\begin{tikzpicture}
	\draw[fill] (0,0)-- (90:#1) arc (90:270:#1) -- cycle ;
	\draw (0,0) circle (#1);
	\end{tikzpicture}}
\newcommand*\fullcirc[1][1ex]{\tikz\fill (0,0) circle (#1);}

\newcommand{\sys}{\texttt{MCPWorld}\xspace}

\newcommand{\ntasks}{\textbf{201}\textsuperscript{*}\xspace}
\newcommand{\napps}{\textbf{8}\xspace}
\newcommand{\cmark}{\textcolor{green}{\ding{51}}} 
\newcommand{\xmark}{\textcolor{red}{\ding{55}}} 

\definecolor{pyblue}{RGB}{0, 0, 150}
\definecolor{pygreen}{RGB}{0, 150, 0}
\definecolor{pyred}{RGB}{150, 0, 0}
\definecolor{pygray}{RGB}{128,128,128}
\title{\sys: A Unified Benchmarking Testbed for\\API, GUI, and Hybrid Computer Use Agents}

\author{%
  Yunhe Yan$^{1,}$$^2$$^*$, Shihe Wang$^1$$^*$, Jiajun Du$^1$$^*$, Yexuan Yang$^1$, Yuxuan Shan$^1$, \\
   \textbf{Qichen Qiu$^1$, Xianqing Jia$^1$, Xinge Wang$^1$, Xin Yuan$^1$, Xu Han$^1$, }
  \\\textbf{Mao Qin$^1$, Yinxiao Chen$^1$, Chen Peng$^1$, Shangguang Wang$^1$, Mengwei Xu$^1$} \\
  \\
  $^1$Beijing University of Posts and Telecommunications, \\
   $^2$Pengcheng Laboratory\\
   $^*$Equal Contribution
}

\begin{document}

\maketitle

\begin{abstract}
  (M)LLM-powered computer use agents (CUA) are emerging as a transformative technique to automate human-computer interaction.
  However, existing CUA benchmarks predominantly target GUI agents, whose evaluation methods are susceptible to UI changes and ignore function interactions exposed by application APIs, e.g., Model Context Protocol (MCP).
  To this end, we propose \sys, the first automatic CUA testbed for API, GUI, and API-GUI hybrid agents.
  A key principle of \sys is the use of ``white-box apps'', i.e., those with source code availability and can be revised/re-compiled as needed (e.g., adding MCP support), with two notable advantages:
  (1) It greatly broadens the design space of CUA, such as what and how the app features to be exposed/extracted as CUA-callable APIs.
  (2) It allows \sys to programmatically verify task completion by directly monitoring application behavior through techniques like dynamic code instrumentation, offering robust, accurate CUA evaluation decoupled from specific agent implementations or UI states.
  Currently, \sys includes 201 well curated and annotated user tasks, covering diversified use cases and difficulty levels.
  \sys is also fully containerized with GPU acceleration support for flexible adoption on different OS/hardware environments.
  Our preliminary experiments, using a representative LLM-powered CUA framework, 
  achieve 75.12\% task completion accuracy, simultaneously
  providing initial evidence on the practical effectiveness of agent automation leveraging MCP.
  Overall, we anticipate \sys to facilitate and standardize the benchmarking of next-generation computer use agents that can leverage rich external tools. Our code and dataset are publicly available at \url{https://github.com/SAAgent/MCPWorld}.
\end{abstract}

\section{Introduction}
\label{sec:introduction}

Computer Use Agents (CUAs)~\citep{computeruse} powered by (multimodal) large language models (LLMs) are rapidly emerging as a transformative technique that augment human productivity. CUAs automate routine tasks, such as document processing, message sending, and application configuration, by understanding natural language instructions and executing them through interaction with computers.

In general, interactions between CUAs and the underlying environment fall into two categories: GUI and API.
A CUA leverages one of them, or a hybrid of both, as observation and action spaces, to accomplish user tasks.
With the standardization of CUA-to-API interface, e.g., model context protocol (MCP)~\citep{mcp_intro}, we anticipate API-powered CUAs to play an increasingly important role in the future.
However, we find that existing benchmarking tools and testbeds mostly target GUI agents only~\citep{osworld,windowsagentarena,androidworld}. 
Unlike CUAs performing solely GUI operations, constructing and benchmarking API-powered CUAs needs to
(i) provide intrusive visibility into app internals and permission to define APIs;
(ii) devise a GUI-independent evaluator that can accurately judge task completion without looking at the device screen.
Currently, only a handful of testbeds provide support for the API CUAs.
Several prior studies~\citep{api_bank, toolbench, toolalpaca, toolqa, taskmatrix} investigated the capability of API CUAs.
But only one recent work, AgentStudio~\citep{agentstudio} supports comparison between two interfaces, but only through simple queries of external output files with no MCP support.

To fill the research gap, we present \textbf{\sys}, the first MCP-enabled benchmarking toolkit and testbed for API, GUI, and hybrid Desktop CUAs.
A key principle of \sys is the use of ``white-box apps'', i.e., those with source code availability and can be revised/re-compiled as needed (e.g., adding MCP support).
Such a design choice not only greatly broadens the design space of CUA, such as what and how the app features to be exposed/extracted as CUA-callable APIs, but also allows \sys to programmatically verify task completion by directly monitoring application behavior through techniques like dynamic code instrumentation, offering robust, accurate CUA evaluation decoupled from specific agent implementations or UI states.
Table~\ref{tab:summary_comparsion} compares \sys to prior CUA testbeds, highlighting its uniqueness in the app selection and evaluation methods.

We conducted considerable effort in building \sys.
At its core, \sys provides a fully containerized desktop environment that ensures reproducible testing conditions, with properly desktop applications setup.
Atop the environment with applications, \sys integrates two sets of unified tools for GUI and MCP interaction.
GUI tool abstracts the functionality of accessing varying GUI representations (e.g., pixel-level screenshots, textual view hierarchies), while leaving the flexibility for customizing app-specific observation spaces.
The MCP tool provide a general framework for direct function-call style interaction with applications; for our white-box applications, we integrated existing MCP servers to expose a rich set of internal operations as standardized APIs for agent use.
For \sys evaluator, we propose and implement three different task verification methods: dynamic binary instrumentation, targeted code injection, and API-driven state querying, all carefully tailored to each application's design. These verification methods synchronize directly with application internal changes, operate independently of the CUA's workflow, and provide immediate feedback on CUA-environment interactions, thus enabling precise, real-time evaluation.

\sys benchmark suite encompasses 201 carefully curated tasks with varying complexity levels across 10 widely-used desktop applications (e.g., VS Code and OBS Studio). 
Tasks are derived and annotated based on realistic user scenarios by 12 authors with computer science backgrounds. 
For each task, we performed in-depth source code analysis, developed custom verification hooks into application internals, and designed application state snapshots to ensure consistent evaluation. 
Tasks range from simple configurations to complex multi-step workflows, with difficulty categorized by human interaction steps and balanced coverage across usage categories.
Each task features detailed natural language instructions and fine-grained progress tracking signals through annotated key milestones, enabling precise evaluation of agent performance.

\sys is designed for scalability and ease of use. We provide standardized workflows and helper tools that simplify the addition of new applications, tasks and toolsets. For agents already supporting the MCP, integration with \sys becomes exceptionally straightforward, allowing researchers to rapidly benchmark their capabilities.

We built a CUA on top of Claude Computer Use~\cite{computeruse} to test its performance on \sys.
Our preliminary results show that Claude Computer Use agent with Claude 3.7 Sonnet achieves 75.12\% task completion accuracy.
We further provide initial evidence on the practical effectiveness of agent automation leveraging MCP. Overall, we anticipate \sys could facilitate and standardize the benchmarking of next-generation CUAs that can leverage rich external tools.

\begin{table}[t]

\centering
\caption{\textbf{Comparing different benchmarks and datasets for computer use agent evaluation.} The meanings of symbols: \emptycirc: all apps used are open-sourced; \fullcirc: all apps used are closed-sourced; \halfcirc: not all apps used are open-sourced; $*$: tasks are provided in the form of templates and can be expanded into different tasks according to different task parameters.
Compared to our novel evaluation method (App internals hooking), we summarize the previous evaluation methods as ``external state matching'', which can be mainly divided into three specific categories: step-wise action matching, external UI matching and output file matching.
}
\label{tab:summary_comparsion}
\begin{adjustbox}{max width=1.0\textwidth}
\begin{tabular}{@{}lcccclcc@{}}
\toprule

\multirow{3}{*}{\textbf{Name}} &
\multicolumn{3}{c}{\textbf{App/Website/Environment}} &
\multirow{3}{*}{\textbf{\# tasks}} &
\multirow{3}{*}{\textbf{Evaluation Method}} &
\multirow{3}{*}{\textbf{Platform}} &
\multirow{3}{*}{\begin{tabular}[c]{@{}c@{}}\textbf{Supported Agent}\\ \textbf{Action Space}\end{tabular}} \\
\cmidrule(lr){2-4}
& 
\begin{tabular}[c]{@{}c@{}}\textbf{Number}\end{tabular} & 
\begin{tabular}[c]{@{}c@{}}\textbf{Open}\\ \textbf{Source}\end{tabular} & 
\begin{tabular}[c]{@{}c@{}}\textbf{MCP}\\ \textbf{Enabled}\end{tabular} & 
& & \\ 
\midrule

AndroidWorld~\citep{androidworld} & 20 & \halfcirc & \xmark & 116\textsuperscript{*} & All three external state matching & Android & GUI \\


AndroidArena~\citep{androidarena} & 13 & \halfcirc & \xmark & 221 & Step-wise action matching & Android & GUI \\


LlamaTouch~\citep{llamatouch} & 57 & \halfcirc & \xmark & 496 & Output file \& external UI matching & Android & GUI  \\

A3~\citep{a3} & 20 & \fullcirc &  \xmark & 201 & Step-wise action \& external UI matching & Android & GUI \\

B-MoCA~\citep{b_moca} & 4 & \fullcirc &  \xmark & 131\textsuperscript{*} & Output file \& external UI matching & Android & GUI \\

\midrule


OSWorld~\citep{osworld} & 9 & \halfcirc  & \xmark & 369 & Output file \& external UI matching & Desktop & GUI \\

WindowsAgentArena~\citep{windowsagentarena} & 11 & \halfcirc & \xmark & 154 & Output file \& external UI matching & Desktop & GUI \\

AgentStudio~\citep{agentstudio} & 9 & \halfcirc  & \xmark & 205 & Output file matching & Desktop & GUI, API, Hybrid \\

AgentBench~\citep{agentbench} & 8 & \halfcirc  & \xmark & 1,091 & Output file \& step-wise action matching & Desktop & GUI \\

VisualAgentBench~\citep{visualagentbench} & 5 & \halfcirc & \xmark & 746 & Output file \& external UI matching  & Desktop & GUI \\

AssistGUI~\citep{assistgui} & 9 & \fullcirc & \xmark & 100 & Output file matching & Desktop & GUI \\


WebArena~\citep{webarena} & 6 & \fullcirc  & \xmark & 241\textsuperscript{*} & Output file \& external UI matching & Web & GUI \\

Visual WebArena~\citep{visualwebarena} & 4 & \fullcirc  & \xmark & 314\textsuperscript{*} & Output file \& external UI matching & Web & GUI \\

WorkArena~\citep{workarena} & 1 & \fullcirc  & \xmark & 33\textsuperscript{*} & Step-wise action matching & Web & GUI \\

WebShop~\citep{webshop} & 1 & \fullcirc  & \xmark & 12k & Step-wise action matching & Web & GUI \\

\midrule

\textbf{MCPWorld} & ~\napps & \emptycirc & \cmark & \ntasks & \textbf{App internals hooking}  & \textbf{Desktop} & \textbf{GUI, API, Hybrid} \\

\bottomrule


\end{tabular}%
\label{tab:compare}
\end{adjustbox}
\end{table}

\section{The \sys Framework}
\label{sec:framework}

\subsection{Framework Overview}
\label{sec:framework_overview}

\begin{figure}[ht]
  \centering
  \includegraphics[width=1.0\linewidth]{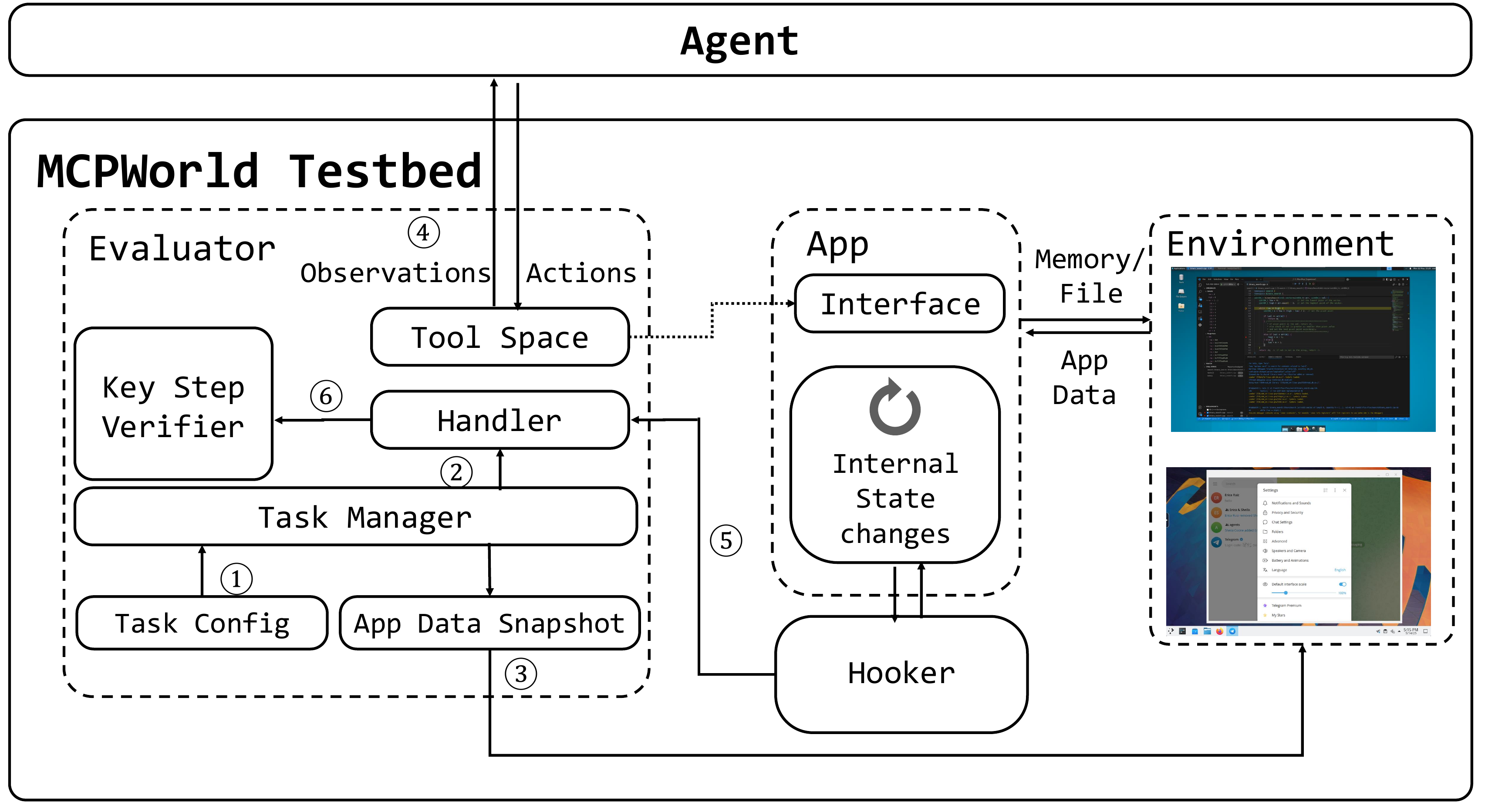}
  \caption{The \sys evaluation workflow. The Task Manager initializes the environment by loading task configurations and application data snapshots. It then starts the application within a Docker container. The Agent interacts with the application through a Unified Tool-based Space, receiving observations and sending actions (GUI or MCP). The Evaluator monitors internal application signals triggered by these interactions and, based on the registered handler, reports task success or failure.}
  \label{fig:architecture}
\end{figure}

\sys provides a unified evaluation platform that supports objective comparison of computer use agent performance across various interaction modalities. Figure~\ref{fig:architecture} illustrates the comprehensive workflow and core components of \sys.~\ding{172}~The process begins with loading the specific task config, which including the task description.
\ding{173}~Register the handler with the app source code to monitor changes in the internal state of the app in real time.
\ding{174}~Load the necessary app data snapshot according to the given task config to establish a consistent starting state, and then launch the target application.
\ding{175}~The agent interacts with the application through the unified tool-based space, which provides observations (e.g., screenshots, UI tree) and accepts actions (e.g. clicks, text input, MCP calls).
\ding{176}~As the agent performs actions, internal state changes within the app trigger signals according to the app source code.
\ding{177}~These signals are captured by the evaluator, which guided by the registered handler from the task configuration, determines the outcome of key steps and ultimately reports overall task success or failure.
This entire orchestrated process, excluding the agent itself, constitutes the \sys Testbed.

Compared to traditional benchmarks, \sys's key advantage lies in its ability to derive verification signals directly from within applications, decoupling evaluation from specific agent implementations or UI states. This approach not only enhances evaluation robustness but also makes comparisons between different interaction strategies (GUI, API/MCP, Hybrid) more fair and meaningful.

\subsection{Tool-based Observation and Action Space}
\label{sec:interaction_modalities}

To enhance agents' ability to interact with both the GUI and API, \sys is designed with a tool-oriented unified observation and action space. Tools are formatted according to the MCP standard and include:
(1) \textbf{GUI tool} enables agents to directly access the basic desktop environment and interact with external devices in a human-like manner. We utilize the GUI tool from Claude Computer Use~\cite{computeruse}, which supports capturing screen images at a specified resolution and performing common mouse and keyboard operations (e.g., clicking, typing, etc.).
(2) \textbf{MCP tool} allows agents to directly interact with predefined APIs, enabling more accurate observation of application status and execution of functions, although the number of available APIs is limited by the application itself.

A unified tool architecture decouples the agent from \sys and offers the following benefits:
(1) It enables convenient control over input modalities. \sys supports three input modes: GUI-only, CLI-only, and multimodal. This setup facilitates research into the agent’s performance across different input modalities.
(2) It provides structured access to GUI information. While raw GUI data is inherently complex and varied, the GUI tool offers a standardized mechanism for requesting and receiving observations such as screenshots or accessibility trees. This abstraction simplifies the agent’s perception module, which can now rely on consistent data formats from the tool instead of parsing diverse raw GUI outputs. Additional GUI representations (e.g., Android’s view hierarchy) can be integrated by extending the capabilities of the MCP server and associated tools.
(3) It allows exisiting app APIs to be integrated in a unified MCP-tool format, offering flexibility for extending functionalities that app exposes in the future.
Instead of being limited to the APIs provided by the testbed, developers can easily extend \sys by adopting a wide range of MCP-formatted community-developed tools across various apps, or simpily design their own. 

\subsection{Environment}
\label{sec:environment} 

The \sys environment aims to provide (1) a standardized task execution environment, and (2) an application context management mechanism for setting up initial app state, such as logging in a test account. 
Traditional testing platforms often grapple with challenges in maintaining precise environmental and state control. To bring the app into a desired state, prior work \citep{windowsagentarena, osworld, agentbench, agentstudio} involves using automation scripts that might modify underlying files. However, for scenarios where internal structures are inaccessible, researchers are often forced to rely on UI automation \citep{llamatouch} to perform the setup process. As a result, this approach is notoriously brittle and time-consuming, susceptible to failures from minor UI alterations. Similarly, while full virtual machine or emulator snapshots~\citep{androidworld} offer comprehensive state capture, their substantial resource footprint and lengthy restoration cycles render them suboptimal for benchmarks requiring frequent and rapid state resets, particularly within a desktop application context.

We use Docker to provide a standardized operating system, consistent application dependencies, and common tooling. Furthermore, a unified, out-of-the-box VNC setup ensures a consistent basic interaction interface for diverse agent modalities.
For application context management, \sys backs up necessary files and restores the application's persistent state when a new task run is initiated.
By ensuring only the application's persistent user data is snapshotted and restored, rather than the entire system image or an application's real-time process state, \sys reduces the setup latency while preserving low system resource cost. 

By providing a consistent environment that supports these diverse interaction modalities for the same application instance, \sys enables controlled studies into the relative strengths, weaknesses, and synergies of GUI-based versus API-centric automation approaches. The subsequent section details the Automated Evaluator, which ensures task success is judged fairly regardless of the interaction modality employed by the agent.

\subsection{Automated Evaluator}
\label{sec:evaluator}

Traditional evaluation methods for CUAs often rely on surface-level observation comparisons. These include matching agent actions against pre-defined UI trajectories~\citep{androidarena, llamatouch, a3, visualagentbench}, inspecting UI accessibility trees~\citep{androidworld, llamatouch, a3, b_moca, osworld, windowsagentarena, visualagentbench, webshop}, comparing resultant files or database states after task completion~\citep{osworld, androidworld, llamatouch, b_moca, windowsagentarena, agentstudio, agentbench, visualagentbench, assistgui, webarena, visualwebarena}, 
While useful in some contexts, these approaches exhibit significant limitations. 

\textbf{External UI matching methods} are especially fragile to interface modifications, suffer from ambiguity in outcome verification and in defining precise checks for diverse tasks. It's also deeply integrated with the GUI interface, ill-suited for API CUAs. 

\textbf{Output file matching methods} (e.g., checking files or databases) also suffer from several drawbacks: 
(1) Representative files or data may be unavailable due to encryption, proprietary formats, or more fundamentally, critical information might exist only in memory and is never persisted to disk at all.
(2) Such methods typically operate synchronously with agent, verifying state only after an agent claims task or single-step completed. This ``after-the-move'' checking makes them ill-suited for verifying intermediate key steps, especially if the crucial information for these steps is transient and not persistently stored. Many critical internal states or momentary UI feedback confirming a sub-step are simply missed.
(3) There can be delays in data persistence, or the recorded state might be vague or incomplete by the time of evaluation, leading to ambiguity in outcome verification.

To address these limitations and robustly support a broader spectrum of agent architectures, including API CUAs, \sys introduces a novel \textbf{``white-box'' evaluation paradigm}. This paradigm shifts evaluation from approaches that synchronize with agent's workflow (might evaluates before an action fully takes effect) and those rely on external, often delayed or incomplete, application outcomes. Instead, it directly responding to \textbf{in-app hook triggering}, intercepts meaningful signals at runtime, aiming to provide accurate, reproducible, and fine-grained assessments.
The core principle is that the most reliable verification of task or intermediate key steps is achieved by identifying and synchronizing with semantically meaningful internal signals within the application.
These signals—including function invocations, event emissions, or internal state transitions—directly and unequivocally correspond to desired outcomes.

\begin{figure}[ht]
  \centering
  \includegraphics[width=1.0\linewidth]{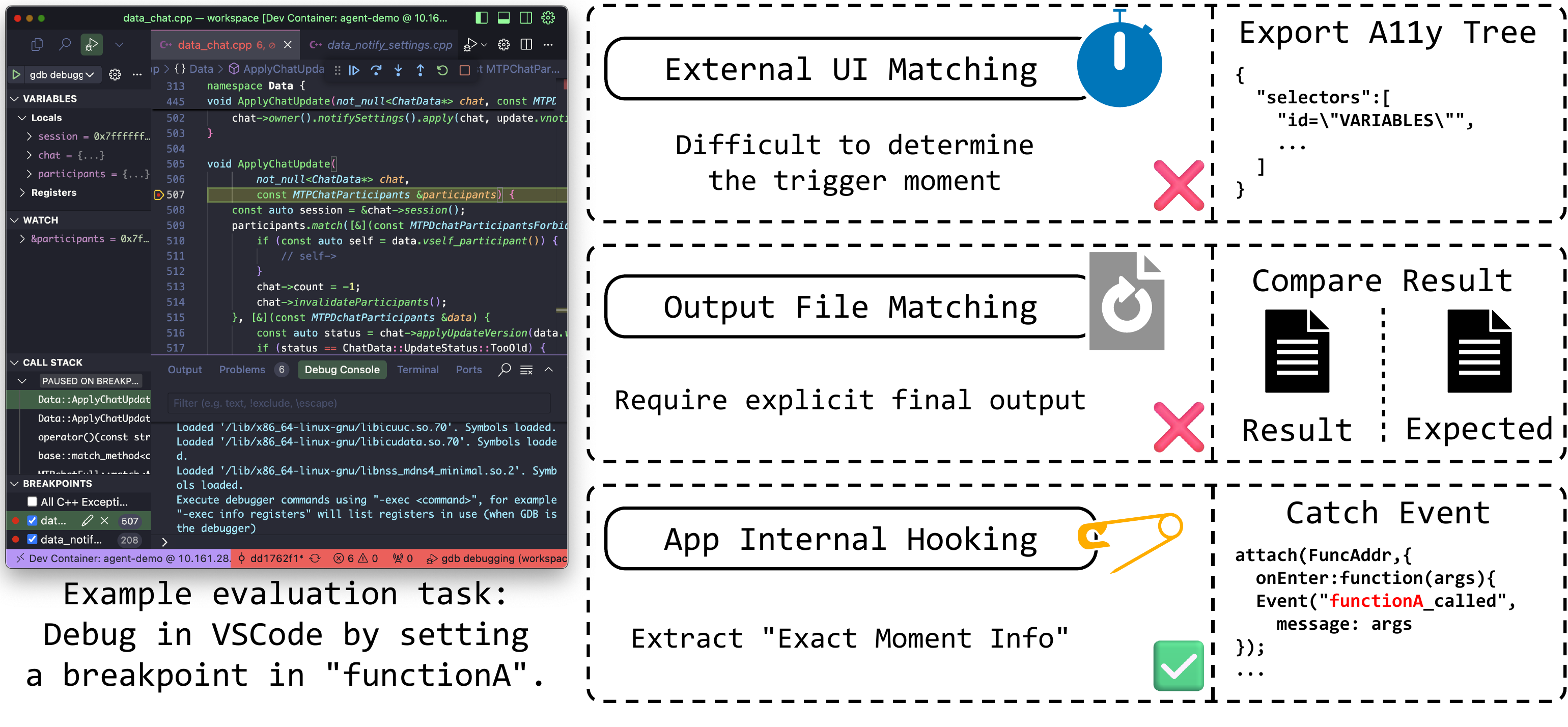}
  \caption{Comparison of evaluation paradigms for a dynamic task: debugging in an IDE. 
  (a) \textbf{External UI Matching}: This approach struggles with timing as the exact moment a breakpoint hits is unpredictable, making it difficult to capture the relevant UI state (e.g., an A11Y tree) for verification. 
  (b) \textbf{Output File Matching}: This method relies on explicit final outputs (e.g., log files, saved documents) to verify task completion. However, crucial in-memory states during debugging (like call stacks or variable values at a breakpoint) are often not persisted, rendering this approach insufficient for verifying such intermediate steps.
  (c) \textbf{App Internal Hooking}: Our approach directly instruments the application to ``Catch Event'' at the exact moment of interest (e.g., a breakpoint hit). It can then extract "Exact Moment Info" such as call stacks and variable states directly from memory, providing robust and precise verification of both intermediate key steps and final outcomes, independent of UI transience or the need for explicit file outputs.
  }
  \label{fig:compare_evaluation_paradigms}
\end{figure}

To further illustrate the advantages of the white-box evaluation paradigm, consider (Figure~\ref{fig:compare_evaluation_paradigms}) an agent performing breakpoint debugging using an IDE (e.g., VS Code). During debugging, the exact moment a breakpoint is triggered depends entirely on runtime branching logic and input data flows, making it inherently unpredictable.

Traditional UI screenshot-based evaluation methods encounter significant limitations in this context. Since we don't know when the exact moment the breakpoint triggers, the evaluator cannot know in advance when to capture a screenshot or export the IDE’s accessibility tree to catch a breakpoint event. Additionally, relying solely on persistent logs or output files for confirmation is similarly unreliable, as crucial debugging states (e.g., call stacks and memory snapshots of variables) exist exclusively in memory during execution, typically not persisted to the file system.

In contrast, our proposed white-box paradigm precisely addresses these issues. Specifically, it can:

\noindent $\bullet$ Utilize hooks inserted into the IDE to respond to internal debugging events in real-time, accurately capturing events the moment a breakpoint is triggered;

\noindent $\bullet$ Immediately extract the call stack and critical variable states at the breakpoint event, reliably determining whether the program halted at the expected location;

\noindent $\bullet$ Verify whether the agent indeed completed the intended debugging steps, independent of fleeting UI feedback or external file-based records.

Thus, the white-box evaluation paradigm exhibits clear and substantial advantages over traditional methods, especially when dealing with highly dynamic, memory-state-dependent tasks such as breakpoint debugging.

\section{The \sys Benchmark Suite}
\label{sec:benchmark_suite}

\subsection{Applications and Tasks}
\label{subsec:benchmark_app_stats}

\begin{table}[t]
  \centering
  \begin{minipage}[t]{0.40\textwidth}\vspace*{0pt}
    \centering
    \captionof{table}{List of \sys\ apps with detailed statistics. All apps have MCP server enabled.}
    \label{tab:selected_apps}
    \begin{adjustbox}{max width=\textwidth}
      \begin{tabular}{@{}lcccc@{}}
        \toprule
        \textbf{Category} & \textbf{App Name} & \textbf{LOC} & \textbf{GitHub Star} & \textbf{\# tasks} \\
        \midrule
        Education   & Zotero             & $4.4\times10^5$ & 11.5k & 16 \\
        Creativity  & OBS Studio         & $8.0\times10^5$ & 64k   & 25 \\
        Productivity& Zulip              & $1.1\times10^6$ & 22.7k & 18 \\
        Productivity& Joplin             & $1.0\times10^6$ & 49.6k & 22 \\
        Development & FreeCAD            & $6.1\times10^6$ & 24.4k & 24 \\
        Education   & QGIS               & $1.3\times10^6$ & 11.4k & 15 \\
        Education   & Anki               & $2.8\times10^5$ & 21.8k & 25 \\
        Development & Visual Studio Code & $2.0\times10^6$ & 171k  & 25 \\
        Productivity& qBittorrent        & $1.1\times10^6$ & 31.5k & 8  \\
        Creativity  & Blender            & $4.8\times10^6$ & 15.4k & 23 \\
        \bottomrule
      \end{tabular}
    \end{adjustbox}

    \vspace{1em}  

    \captionof{table}{Step count distribution for tasks in the \sys\ dataset.}
    \label{table:dataset-stats}
    \begin{tabular}{lcr}
      \toprule
      GUI Steps & \# tasks & Difficulty Level \\
      \midrule
      0-5   & 73 & Easy   \\
      5-10  & 83 & Medium \\
      10+   & 45  & Hard   \\
      \bottomrule
    \end{tabular}
  \end{minipage}%
  \hfill
  \begin{minipage}[t]{0.48\textwidth}\vspace*{0pt}
    \centering
    \includegraphics[width=\linewidth]{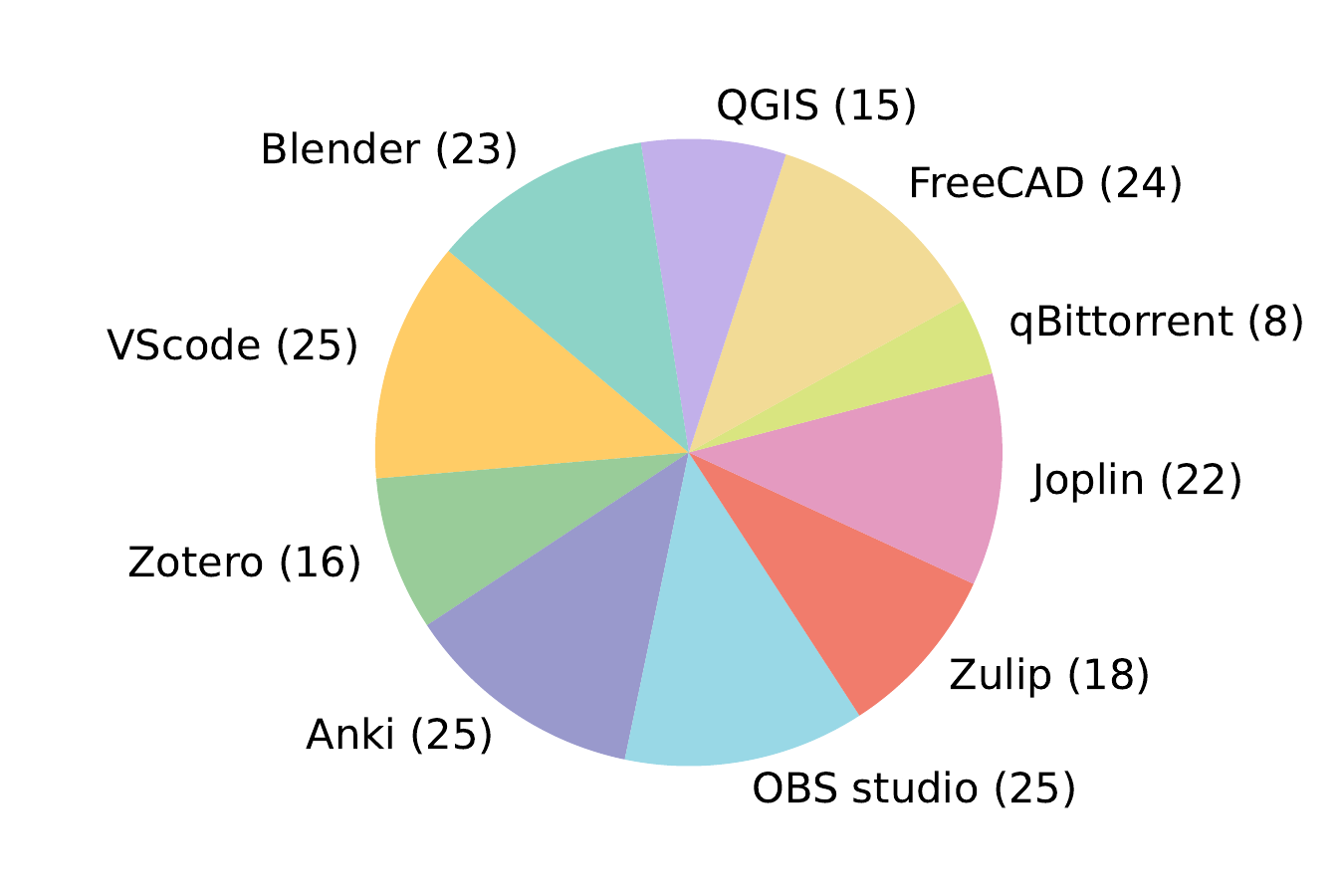}
    \captionof{figure}{Distribution of tasks in \sys\ spanning ten open source applications.}
    \label{fig:task-distribution}
  \end{minipage}
\end{table}

The current \sys benchmark suite includes 10 applications across diverse categories, as shown in Table~\ref{tab:selected_apps}.
These applications cover a broad range of everyday desktop usage scenarios, from code editors and browsers to media players and productivity tools, providing a comprehensive test of agent capabilities. All applications are open-sourced, allowing us to implement our white-box evaluation approaches. When selecting apps, we specifically considered apps that had an MCP Server (either provided by the official developers or third-party community). By integrating our own version of an MCP client, we enable agents to connect to all these MCP servers in a manner similar to the Claude Desktop~\citep{claudedesktop} to obtain MCP capability.

\sys contains 201 tasks distributed across the above applications.
Tasks are designed to mirror realistic user scenarios, ranging from simple actions to complex workflows.

In designing tasks, our core specifications were: (1) covering the app's fundamental functionalities, and (2) establishing reasonable initial states (pre-configured test accounts, opened workspaces and files in VSCode, etc.) as the starting point for task progression. Following these principles, we developed tasks through a combination of human choices, LLM suggestions, and technical documentation, blogs, and forums related to the applications. After designers developed task instructions and related files based on real-world materials and issues, each example underwent cross-validations by two additional task designers who then verified feasibility, clarity, and consistency with source materials.

For each task, annotators recorded themselves completing the task using the DuckTrack~\citep{ducktrack} tool, ensuring each task can be done by human. We instructed annotators to interact naturally with devices, avoid interacting with irrelevant content, and minimize unnecessary scrolling during recordings.

Tasks are also categorized by difficulty. Task difficulty is estimated based on human GUI step count, with easy tasks typically requiring 1-5 steps, medium tasks requiring 5-10 steps, and hard tasks requiring 10+ steps. The distribution of task types is balanced, covering configuration changes, content creation, information retrieval, media manipulation, and other scenarios, ensuring a comprehensive assessment of agent capabilities.

Most tasks feature annotated intermediate milestones (key steps), providing support for fine-grained assessment.
Table~\ref{table:dataset-stats} shows the distribution of task difficulties across applications. These difficulties are estimated based on human GUI interaction step counts, which we derive from human-executed traces. This process ensure that each application includes tasks of varying difficulty levels.

An overview of the statistics for the \sys benchmark suite is shown in Table~\ref{fig:task-distribution}.

\subsection{Implementing Task Verification Logic}
\label{subsec:benchmark_verification_impl}

For each task, annotators first preliminarily analyzed potential verification logic locations in source code repositories (LLMs proved particularly helpful here). Through debugging analysis, annotators then determined the correct execution logic, including call stack relationships and data flows. From such logic, annotators identified the most critical moments representing definitive completion stages (e.g., functions updating memory after receiving remote server update). Finally, annotators base on all state information during correct execution (function parameters, stack frame conditions, return values) to implement robust verification logic.

Once key internal state changes were identified, under the open-source nature of the applications, the core principle in Section~\ref{sec:evaluator} is realized using the most suitable verification technique for the specific application design. We implement the following techniques to access applications' internal signals.

\textbf{Dynamic Instrumentation.} 
For compiled applications (e.g., programmed using C++, Rust with Qt/GTK), we use tools like Frida~\citep{frida} to enable dynamic instrumentation. Frida is helpful for intercepting function calls, accessing internal memory, and injecting custom scripts for verification logic without source-code modification. The verification logic hooks the precise function identified as the ground truth signal for a key step or the final outcome (e.g., the internal API call for adding an item to a cart - key step; the call confirming order placement - final success). The hook validates context to ensure correctness.
	
\textbf{Targeted Code Injection.}
For applications built with more introspective languages (e.g., Python, JavaScript/Electron, Java), we employ targeted code injection.
It introduces small, specialized internal probes into the application's own environment that leverage native runtime capabilities to access internal objects directly, subscribe to events, or execute verification logic. The injected verification logic listens for the specific internal event or monitors the state change confirming successful completion of a key step (e.g., a "file compiled successfully" event) or the overall task.

 \textbf{API-Driven State Querying.}
When applications expose states via APIs (internal, diagnostic, or CLI) or record events in structured logs/databases, \sys proactively queries these APIs to retrieve state or parse logs/databases to verify both intermediate milestones (e.g., a temporary file created) and the final goal (e.g., the correct final entry in a database).

These techniques enable automated evaluation across a wide range of software and ensure that the evaluation process remains decoupled from the software itself.

\section{Benchmarking GUI, MCP, and Hybrid Computer Use Agents}
\label{sec:experiments}

To demonstrate the utility of \sys and gain empirical insights into the effectiveness of different interaction modalities, we evaluate a representative LLM-powered CUA.
The primary goal was to leverage the controlled environment and agent-agnostic evaluation provided by \sys to systematically compare GUI and API strategies within the same agent framework. All experiments were conducted on a Linux server with 64-core CPU and 8 NVIDIA A40 GPUs. \sys's environment is implemented to be compatible with GPU acceleration, but it is worth noting that our experiments do not require significant computational resources and many tasks can be performed without GPU acceleration.

\subsection{Experimental Setup}
\label{subsec:exp_setup}

\paragraph{CUA}
We utilized the official Claude \texttt{computer-use-demo}~\citep{computeruse} framework as our base agent. For all experiments, we used Anthropic's \texttt{claude-3-7-sonnet-20250219} model via its official API, employing a standard ReAct-style~\citep{yao2023reactsynergizingreasoningacting} prompting strategy for planning and tool execution. Specific agent parameters like maximum steps per turn were kept consistent across all runs.

\paragraph{Tools}
The core of our experimental design lies in varying the toolset available to the CUA across three distinct configurations:

\noindent $\bullet$ \textit{GUI-Only.} The agent was restricted to using only standard GUI interaction tools. This included observation tools like screen capture, low-level keyboard and mouse emulation (via \texttt{xdotool}), and executing bash commands.

\noindent $\bullet$ \textit{MCP-Only.} For this configuration, we augmented the CUA framework by our own MCP clients. Each MCP tool corresponds to a specific function exposed by the application's MCP server. The agent in this setup could \emph{only} use these MCP tools, forcing reliance on the programmatic interface.

\noindent $\bullet$ \textit{Hybrid.} The agent was provided with the \emph{union} of the GUI-Only and MCP-Only toolsets. The LLM selects the most appropriate tool(s) for each step based on the task requirements. No explicit heuristics were used to bias tool selection.

All experiments were conducted on the full \sys benchmark suite, encompassing 10 applications and 201 tasks.
Each task was executed within the standard \sys containerized environment to ensure consistency. Each task within the benchmark was attempted 3 times for each of the three tool configurations (GUI-Only, MCP-Only, Hybrid). Task success was determined automatically by the \sys Evaluator using the pre-defined verification logic for final goals and key steps, ensuring an objective and modality-agnostic assessment. A maximum time limit of 300 seconds was enforced per task attempt.

\paragraph{Metrics}
We evaluate performance using the following primary metrics, automatically computed by \sys:

\noindent $\bullet$ \textit{Task Success Rate (SR).} The percentage of runs where the agent successfully achieved the final task goal within the time limit of 300 seconds.

\noindent $\bullet$ \textit{Key Step Completion Rate (KSCR).} The 
percentage of annotated key steps successfully completed per run. This provides a measure of partial progress, especially valuable for complex tasks. 

\subsection{Results}
\label{subsec:exp_results}

\begin{table}[ht]
\centering
\caption{Overall performance comparison across interaction modalities (averaged over all tasks and runs). }
\label{tab:overall_results}
\begin{adjustbox}{max width=0.8\textwidth}
\begin{tabular}{lcc}
\toprule
Configuration & Task Success Rate (\%) & Key Step Completion Rate (\%) \\
\midrule
GUI-Only      & 70.65                   & 68.82                          \\
MCP-Only      & 53.23                   & 59.78                          \\
Hybrid        & 75.12                   & 69.63                          \\
\bottomrule
\end{tabular}
\end{adjustbox}
\end{table}

Our experiments demonstrated distinct performance differences among the three interaction modalities. Table~\ref{tab:overall_results} presents the overall performance averaged across all tasks in the benchmark. The Hybrid configuration achieved the highest Task Success Rate at 75.12\%, outperforming both the GUI-Only (70.65\%) and MCP-Only (53.23\%) configurations. Additionally, the Hybrid setup maintained a strong Key Step Completion Rate (69.63\%), closely comparable to the GUI-Only configuration (68.82\%), which surpassed the MCP-Only configuration (59.78\%). The MCP-Only configuration consistently underperformed compared to both GUI-Only and Hybrid setups in terms of both Task Success Rate and Key Step Completion Rate.

\subsection{Analysis}
\label{subsec:exp_analysis}

\begin{table}[ht]
\centering
\caption{Distribution of task failure reasons across different interaction configurations (\% of total failures).}
\label{tab:failure_reasons}
\begin{adjustbox}{max width=1.0\textwidth}
\begin{tabular}{lccc}
\toprule
\textbf{Failure Reason} & \textbf{GUI-Only} & \textbf{MCP-Only} & \textbf{Hybrid} \\
\midrule
Imprecise cursor positioning        & 5.08\%    & 0.00\%    & 2.08\% \\
UI element not found                & 1.69\%    & 0.00\%    & 0.00\% \\
Limited reasoning capability        & 72.88\%   & 47.37\%   & 85.41\% \\
MCP API incompatibility             & 0.00\%    & 3.16\%   & 0.00\% \\
Insufficient MCP coverage           & 0.00\%    & 48.42\%   & 4.17\% \\
Insufficient MCP tool description   & 0.00\%    & 1.05\%    & 2.08\% \\
Time limit exceeded                 & 18.64\%    & 0.00\%    & 2.08\% \\
Others                              & 1.69\%    &   0.00\%   &  4.17\% \\
\bottomrule
\end{tabular}
\end{adjustbox}
\end{table}

\begin{table}[ht]
\centering
\caption{Success rates of GUI, API, and Hybrid configurations for tasks of different complexity.}
\label{tab:succ_step_diff}
\begin{adjustbox}{max width=1.0\textwidth}
\begin{tabular}{lccc}
\toprule
\textbf{GUI steps} & \textbf{GUI-Only} & \textbf{MCP-Only} & \textbf{Hybrid} \\
\midrule
0-5        & 90.41\%    & 63.01\%    & 90.41\% \\
5-10        & 72.29\%    & 51.81\%    & 74.70\% \\
10+       &   35.56\%     & 40.00\%   & 51.11\% \\
\bottomrule
\end{tabular}
\end{adjustbox}
\end{table}

The most notable comparison result is the low success rate of MCP-only agents. As indicated by the failure attribution in Table~\ref{tab:failure_reasons}, this is largely due to limited MCP functionality and coverage, which also limits the Hybrid CUA’s improvement over the GUI-only baseline.

We then analyzed task success rates across different difficulty levels. As task complexity increases, the GUI-only CUA’s success rate drops by 54.85\%, while MCP-only and Hybrid CUA decline by only 23.01\% and 39.30\%, respectively. This indicates: (1) MCP support improves robustness by enabling more direct, deterministic control and reducing reliance on error-prone GUI interactions—especially for complex tasks; (2) Hybrid CUA benefits from combining both paradigms, using MCP when APIs are available and falling back to GUI when needed, resulting in significantly enhanced resilience and robustness.

In general, enabling MCP enhances the capabilities of CUA. However, as shown in Table~\ref{tab:succ_step_diff}, the success rate dropped after adding MCP support for tasks comprising 5-10 steps. Our preliminary analysis suggests this decline might due to interruption made by MCP tools' overly long prompt, which increased the likelihood of mistakes that GUI-only CUA won't make.

Our experimental results show substantially higher task success rates compared to prior work~\citep{osworld, a3, androidworld, windowsagentarena, visualagentbench}. We attribute this to three main factors: (1) a more flexible action space that provices CUAs more freedom in decision-making, such as executing command line instructions; (2) the ability to easily recover from erroneous actions, enabling repeated attempts; and (3) a task set primarily composed of common, everyday tasks. In future work, we plan to introduce more complex tasks, including those involving interactions across multiple applications.

\section{Limitations}

While \sys represents a significant advancement in benchmarking CUAs, we acknowledge several limitations. First, our current preliminary experiments focus on a single agent framework, and future work will incorporate a broader range of agent architectures to enable more comprehensive comparative analyses. Second, our reliance on white-box, open-source applications limits the breadth of software that can be included in the benchmark. However, this limitation is by design and is intended to support more robust and in-depth evaluations. Third, task annotations require a deeper understanding of the application source code, though we have found that LLMs can significantly reduce this burden. Finally, our current benchmark suite would benefit from further expansion to include more apps and tasks, as well as complex scenarios like cross-application workflows and multi-turn interactions. Despite these limitations, \sys provides a robust foundation for standardized CUA benchmarking that can evolve alongside advancements in the field.

\section{Conclusion}
\label{sec:conclusion}

We introduced \sys, a novel, unified testbed enabling robust white-box evaluation of API, GUI, and hybrid CUAs by directly hooking internal application behavior across its current suite of approximately 201 tasks and 10 applications. Our preliminary experiments, while indicating the potential of hybrid approaches, reveal significant challenges for current agents in tackling complex real-world tasks. Future work will focus on expanding the task and application library, designing more intricate task mechanisms such as multi-turn scenarios and complex cross-application workflows, and testing a wider variety of agent architectures. We believe these efforts will accelerate the development of more capable and reliable CUAs, ultimately fostering more intelligent human-computer interaction.

\section{Acknowledgements}
This work was supported in part by the Major Key Project of PCL under Grant PCL2024A06 and PCL2022A05, and in part by the Shenzhen Science and Technology Program under Grant RCJC20231211085918010.

\bibliographystyle{plain}

\nocite{*}

\newpage

\appendix
\section*{Appendix} 

\section{Details of the \sys Environment}
\label{app:environment_details}

\subsection{Overall Infrastructure}
\label{app:overall_infrastructure}
The \sys execution environment is built upon an Ubuntu 22.04 container. We provide two primary desktop environment configurations to cater to diverse experimental needs and application requirements:

\noindent $\bullet$ \textbf{Lightweight XFCE Environment:} This configuration pairs the XFCE desktop environment with TurboVNC. XFCE is chosen for its minimal resource footprint and simplicity, while TurboVNC provides efficient remote desktop access. This setup is ideal for tasks that do not require extensive graphical capabilities or GPU acceleration, offering a clean and responsive environment for general application interaction.

\noindent $\bullet$ \textbf{Full-Featured KDE Environment with GPU Acceleration:} For more demanding tasks, particularly those benefiting from GPU hardware acceleration or requiring a more comprehensive desktop experience, we offer a configuration based on the KDE Plasma desktop. This environment utilizes KasmVNC for remote access and integrates technologies from the Selkies project to enable robust GPU acceleration pass-through to the container. This allows agents to interact with applications that leverage GPU capabilities (e.g., for rendering, media processing, or AI-assisted features) within a rich desktop environment.

Both environments are designed to provide a stable and isolated platform for CUA evaluation. The choice of environment can be configured on a per-task or per-experiment basis, allowing researchers to select the most appropriate setup for their specific investigation. This dual-environment approach ensures that \sys can support a wide range of applications and agent interaction scenarios, from simple UI manipulations to complex, graphically intensive workflows.

\subsection{Details of Tool-based Observation and Action Space}
\label{app:tool_space_details}

The interaction between the agent and the \sys environment is mediated through a unified, tool-based paradigm. This design, which forms the basis of our current reference agent implementation, is adapted from the Anthropic \texttt{computer-use-demo} framework. While our specific agent leverages this existing tooling, the \sys testbed and its white-box evaluation methodology are designed to be compatible with other agent architectures that can interact via a similar tool-use abstraction in the future.

\paragraph{Observation Space}
The agent gathers information about the environment by calling observation-oriented tools, summarized in Table~\ref{tab:observation_tools}.

\begin{table}[h!]
\centering
\caption{Detailed Observation Tools in \sys}
\label{tab:observation_tools}
\begin{tabularx}{\linewidth}{@{}l l >{\RaggedRight\arraybackslash}X@{}}
\toprule
Tool Category & Specific Action & Description \\
\midrule
\multirow{2}{*}{\texttt{ComputerTool}} & \texttt{screenshot} & Captures the entire display. The screenshot is returned to the agent as a Base64-encoded PNG image. \\
                                     & \texttt{cursor\_position} & Retrieves the current (x, y) coordinates of the mouse cursor. \\
\midrule
\texttt{BashTool} & (N/A - General Purpose) & Executes read-only shell commands (e.g., \texttt{cat <filename>}, \texttt{grep}) to retrieve file content or command output. \\
\midrule
\texttt{EditTool} & \texttt{view} & Inspects and retrieves content from text strings or files. \\
\midrule
MCP-driven Tools & (Dynamically Discovered) & Provide structured observations directly from applications by invoking read-only MCP tools exposed by an MCP server. \\
\bottomrule
\end{tabularx}
\end{table}

\paragraph{Action Space}
The agent performs actions within the environment by invoking action-oriented tools, summarized in Table~\ref{tab:action_tools}. The interaction with each tool follows a standardized format, as managed by the agent's main interaction loop.

\begin{table}[h!]
\centering
\caption{Detailed Action Tools in \sys}
\label{tab:action_tools}
\begin{tabularx}{\linewidth}{@{}l l >{\RaggedRight\arraybackslash}X@{}}
\toprule
Tool Category & Specific Action & Description \\
\midrule
\multirow{14}{*}{\texttt{ComputerTool}} 
& \texttt{key} & Presses a single specified key (e.g., 'enter', 'esc', 'a', 'Ctrl+c'). \\
& \texttt{type} & Types the provided string of characters at the current cursor position or active input field. \\
& \texttt{mouse\_move} & Moves the mouse cursor to the specified (x, y) coordinates on the screen. \\
& \texttt{left\_click} & Performs a left mouse click at the current cursor position or specified coordinates. \\
& \texttt{left\_click\_drag} & Performs a left mouse click at a starting (x,y) coordinate, drags the cursor to an ending (x,y) coordinate, and then releases the mouse button. \\
& \texttt{right\_click} & Performs a right mouse click at the current cursor position or specified coordinates. \\
& \texttt{middle\_click} & Performs a middle mouse click at the current cursor position or specified coordinates. \\
& \texttt{double\_click} & Performs a double left mouse click at the current cursor position or specified coordinates. \\
& \texttt{left\_mouse\_down} & Presses and holds the left mouse button down at the current cursor position or specified coordinates. \\
& \texttt{left\_mouse\_up} & Releases the left mouse button, typically after a \texttt{left\_mouse\_down} or drag operation. \\
& \texttt{scroll} & Scrolls the mouse wheel (e.g., 'up', 'down', or by a specified amount). \\
& \texttt{hold\_key} & Presses and holds a specified modifier key (e.g., 'ctrl', 'shift', 'alt'). The key remains held until explicitly released or until the next tool call. \\
& \texttt{wait} & Pauses agent execution for a specified duration (e.g., in seconds). \\
& \texttt{triple\_click} & Performs a triple left mouse click at the current cursor position or specified coordinates. \\
\midrule
\texttt{BashTool} & (N/A - General Purpose) & Allows the agent to execute arbitrary shell commands within the environment's terminal (e.g., for file system manipulation, application launching, process management, system configuration). \\
\midrule
\multirow{4}{*}{\texttt{EditTool}} 
& \texttt{create} & Creates a new file with specified content, or overwrites an existing file. \\
& \texttt{str\_replace} & Performs string find-and-replace operations within a given text string or the content of a specified file. \\
& \texttt{insert} & Inserts a specified text string into a file at a given line number or appends to the end of the file. \\
& \texttt{undo\_edit} & Undoes the last modification made by the EditTool, if supported by the tool's internal state. \\
\midrule
MCP-driven Tools & (Dynamically Discovered) & Execute application-specific functionalities by invoking MCP tools exposed by an MCP server (e.g., sending a message in a chat application, creating a new document with specific content, triggering an internal application function). \\
\bottomrule
\end{tabularx}
\end{table}

When an agent decides to use a tool, it formulates a request, typically as a JSON object, adhering to the tool's specific \texttt{input\_schema}. This schema defines the parameters the tool accepts. Upon execution, a tool returns a \texttt{ToolResult} object. This object encapsulates:
\texttt{output}, A string containing textual output from the tool (e.g., result of a bash command);
\texttt{base64\_image}, A Base64 encoded string representing an image, typically for screenshot results;
\texttt{error}, A string describing an error if the tool execution failed;
\texttt{system} (optional), A string providing auxiliary system-level feedback or context to the agent related to the tool's execution.
The main loop function formats this \texttt{ToolResult} into a structure that is then passed back to the LLM.

\paragraph{Compatibility with MCP as a Tool}
A key feature of this architecture is its seamless integration of MCP-provided functionalities as standard tools. During initialization, the agent framework connects to configured MCP servers and retrieves a list of available MCP tools. These MCP tools, along with their descriptions and input schemas, are dynamically added to the agent's list of usable tools.

From the agent's perspective, invoking an MCP tool is identical to invoking any other locally implemented tool (like \texttt{ComputerTool} or \texttt{BashTool}). The agent uses the tool's name and provides the required input parameters. The underlying \texttt{MCPClient} (detailed in Appendix~\ref{app:mcp_client_implementation}) handles the communication with the respective MCP server to execute the tool and retrieve the result. This abstraction allows for a unified action and observation space, simplifying agent development and enabling flexible extension with new capabilities exposed via MCP.

\subsection{Implementation of MCP Client}
\label{app:mcp_client_implementation}
To facilitate agent interaction with MCP-enabled applications, we implemented an \texttt{MCPClient}. This client is designed to enable our agent to operate similarly to Claude Desktop, capable of leveraging tools exposed by various MCP servers. The key functionalities provided by our \texttt{MCPClient} include:

\noindent $\bullet$ \textbf{Dynamic Connection to MCP Servers:} The client can establish and manage connections to one or more MCP servers based on provided configurations (e.g., server command and arguments). It utilizes the standard MCP library for communication, typically over stdio.

\noindent $\bullet$ \textbf{Automated Tool Discovery:} Upon connecting to an MCP server, the client automatically discovers the tools offered by that server. It retrieves tool names, descriptions, and their input schemas.

\noindent $\bullet$ \textbf{Standardized Tool Representation:} Discovered MCP tools are presented to the agent in the same format as locally defined tools (as \texttt{anthropic.types.beta.BetaToolParam}), allowing for uniform integration into the agent's decision-making process.

\noindent $\bullet$ \textbf{Transparent Tool Invocation:} The client handles the invocation of MCP tools requested by the agent. It directs the call to the appropriate MCP server and manages the underlying communication.

\noindent $\bullet$ \textbf{Consistent Result Formatting:} Results from MCP tool executions (whether text or image data) are packaged into a standard \texttt{ToolResult} object, identical in structure to results from local tools. This ensures consistent processing of tool outputs by the agent.

\noindent $\bullet$ \textbf{Resource Management:} The client ensures proper cleanup and release of resources associated with MCP sessions upon completion of interactions.

In essence, the \texttt{MCPClient} acts as a bridge, abstracting the specifics of MCP communication and allowing the agent to seamlessly incorporate a wide array of external application functionalities into its operational toolkit.

To facilitate agent interaction via MCP, \sys primarily integrates existing MCP server implementations for the selected open-source applications. Many of these MCP servers are community-driven or third-party projects, often available on platforms like GitHub, and thus exhibit varying levels of feature coverage and maturity. The diversity in available MCP functionalities reflects the current state of MCP adoption and development within the broader open-source ecosystem.

The integration of these MCP servers into the \sys framework is streamlined by our \texttt{MCPClient}. The MCP server configuration specifies how its corresponding  program should be launched. This typically includes the command to execute the server, any necessary arguments, and required environment variables.

During the setup phase for a task that involves MCP interaction, \sys automatically initiates the configured MCP server process for the relevant application. The \texttt{MCPClient} then connects to this server, allowing the agent to discover and utilize the MCP tools exposed by that specific server instance. This declarative approach to enabling MCP support simplifies the addition of new MCP-enabled applications to the benchmark and allows \sys to readily leverage ongoing community efforts in developing MCP servers.

\section{Application and Initial State Setup}
\label{app:initial_state_setup}

\subsection{Binding Application-Under-Test}
\label{app:bindinng}
A core tenet of \sys is the use of open-source applications, which allows for their compilation directly from source code. This control over the build process is essential for in-depth analysis and the implementation of our white-box evaluation methodologies.

Once an application is compiled, its executable, along with any necessary runtime files, is dynamically integrated into the \sys Dockerized environment using bind mounts. This technique establishes a direct link between a directory on the host system (or a build volume) containing the compiled application artifacts and a designated target path within the Docker container. The configuration for such a mount typically specifies the source path on the host, the target path within the container, and the mount type, for example:
\begin{verbatim}
 "source=<path_to_executable_on_host>,target=<path_in_container>,type=bind"
\end{verbatim}
This strategy of dynamically binding pre-compiled applications into the testing environment offers significant advantages. It allows for the flexible substitution of different application versions or custom builds (e.g., for debugging or with specialized instrumentation) without necessitating a rebuild of the base Docker image. This ensures an agile development workflow and a consistently configured application environment for reproducible benchmarking.

\subsection{User Data and Initial State Restoration}
\label{app:user_data_restoration}
To ensure consistent and reproducible experimental conditions for each task run, \sys implements a mechanism for managing and restoring application-specific user data and initial states. Many tasks require applications to be in a particular state before the agent begins interaction, such as being logged into a specific account, having certain files or projects open, or particular configurations applied.

Our approach is to allow each task to define its own prerequisite context data. As illustrated in task configuration files (e.g., \texttt{config.json} for a given task), a task can specify a list of data sources and their corresponding target locations within the evaluation environment's file system. For instance, a task might define:
\begin{verbatim}
"context_data": [
    { "from": "path/to/task_specific_user_profile", 
        "to": "/root/.config/AppName" },
    { "from": "path/to/task_project_files", 
        "to": "/workspace/project_X" }
]
\end{verbatim}
During the initialization phase of an environment for a specific task, \sys processes these definitions. A script is responsible for this restoration process. It iterates through the specified context data entries and, for each entry, copies the contents from the defined source path (``from'') to the target path (``to'') within the container. This operation is typically performed using a robust file synchronization utility like \texttt{rsync} with options ensuring that the target directory accurately mirrors the source (e.g., \texttt{rsync -av --delete source/ target}).

This method of task-specific data restoration guarantees that every execution of a task starts from an identical and well-defined application state, isolating the evaluation of the agent's performance from variations in prior interactions or environmental drift. It also simplifies the creation of diverse tasks, as each can maintain its own tailored initial conditions without interference.

\section{Experimental Setup and Evaluation Configuration}
\label{app:evaluation_configuration}

\subsection{Configuration and Hyperparameters}
\label{app:hyperparameters}
All experiments were conducted using a standardized agent configuration, executed via the same script. The key parameters influencing the agent's behavior and ensuring consistency and reproducibility across experimental runs are detailed below.

\noindent $\bullet$ \textbf{Model:} Anthropic's \texttt{claude-3-7-sonnet-20250219} model served as the core foundation model.

\noindent $\bullet$ \textbf{Temperature:} The sampling temperature was set to 0. This setting minimizes randomness in the LLM's responses, aiming for the most deterministic and probable output at each generation step. This is crucial for enhancing the reproducibility of agent behavior across identical task runs.

\noindent $\bullet$ \textbf{Tool Version:} The agent utilized and enhanced the \texttt{computer\_use\_20250124} version of the Claude toolset, defining the available local tools and their functionalities.

\noindent $\bullet$ \textbf{Timeout}: A global timeout of 300 seconds was enforced for the completion of each task attempt.

\noindent $\bullet$ \textbf{Interaction Modality Configuration:}
The agent's available toolset was systematically varied to evaluate different interaction strategies. This was primarily managed by configuring which categories of tools were accessible to the agent in distinct experimental setups:
    \begin{itemize}
        \item \textbf{GUI-Only Setup:} In this configuration, the agent had access to the \texttt{ComputerTool} for screen interaction (observations like screenshots, and actions like mouse clicks and keyboard inputs), the \texttt{BashTool} for command-line operations, and the \texttt{EditTool} for viewing and modifying file content. MCP-driven tools were disabled.
        \item \textbf{MCP-Only Setup:} Here, the agent could only utilize tools dynamically discovered and provided via the \texttt{MCPClient} from connected MCP servers. The \texttt{ComputerTool} (for direct GUI manipulation) was disabled. The \texttt{BashTool} and the \texttt{EditTool} might still be available depending on the specific experimental variant (see BASH ablation study).
        \item \textbf{Hybrid Setup:} This configuration provided the agent with the union of tools available in the GUI-Only and MCP-Only setups. The agent could leverage the \texttt{ComputerTool}, the \texttt{BashTool}, the \texttt{EditTool}, and any available MCP-driven tools.
        \item \textbf{No-BASH Variants:} For specific ablation studies (as detailed in Appendix~\ref{app:bash_impact_analysis}), the \texttt{BashTool} and \texttt{EditTool} was explicitly disabled across the GUI-Only, MCP-Only, and Hybrid setups to assess its impact. In these variants, the prompt was also adjusted accordingly to reflect the absence of shell command capabilities.
    \end{itemize}

\subsection{Prompt Templates}
\label{app:prompt_templates}
The agent's interaction with the LLM is guided by system-level instructions, task-specific inputs, and dynamically generated information about available tools. Our reference agent, adapted from the Anthropic \texttt{computer-use-demo} framework, primarily utilizes a ReAct-style prompting strategy. The core configurable components related to prompting in our experiments are the system prompts.

\paragraph{Core System Prompts}
A foundational element is the core system prompt, providing the LLM with essential context about its operational environment and general behavioral guidelines. This prompt is explicitly defined within the agent's codebase. Depending on the specific experimental configuration (e.g., GUI-only, MCP-only, Hybrid, and presence or absence of the BASH tool), different versions of the system prompt were employed. The primary system prompts used are detailed below:

\noindent \textbf{1. Default System Prompt:}
Used in standard configurations where GUI interaction, BASH tool, and potentially MCP tools are available.
\begin{lstlisting}[caption=Default System Prompt, basicstyle=\ttfamily\tiny]
SYSTEM_PROMPT = f"""<SYSTEM_CAPABILITY>
* You are utilising an Ubuntu virtual machine using {platform.machine()} architecture with internet access.
* You can feel free to install Ubuntu applications with your bash tool. Use curl instead of wget.
* To open firefox, please just click on the firefox icon.  Note, firefox-esr is what is installed on your system.
* Using bash tool you can start GUI applications, but you need to set export DISPLAY={os.getenv("DISPLAY")} and use a subshell. For example "(DISPLAY={os.getenv("DISPLAY")} xterm &)". GUI apps run with bash tool will appear within your desktop environment, but they may take some time to appear. Take a screenshot to confirm it did.
* When using your bash tool with commands that are expected to output very large quantities of text, redirect into a tmp file and use str_replace_editor or `grep -n -B <lines before> -A <lines after> <query> <filename>` to confirm output.
* When viewing a page it can be helpful to zoom out so that you can see everything on the page.  Either that, or make sure you scroll down to see everything before deciding something isn't available.
* When using your computer function calls, they take a while to run and send back to you.  Where possible/feasible, try to chain multiple of these calls all into one function calls request.
* The current date is {datetime.today().strftime('%A, %B %-d, %Y')}.
</SYSTEM_CAPABILITY>

<IMPORTANT>
* When using Firefox, if a startup wizard appears, IGNORE IT.  Do not even click "skip this step".  Instead, click on the address bar where it says "Search or enter address", and enter the appropriate search term or URL there.
* If the item you are looking at is a pdf, if after taking a single screenshot of the pdf it seems that you want to read the entire document instead of trying to continue to read the pdf from your screenshots + navigation, determine the URL, use curl to download the pdf, install and use pdftotext to convert it to a text file, and then read that text file directly with your StrReplaceEditTool.
</IMPORTANT>"""
\end{lstlisting}

\noindent \textbf{2. API-Only System Prompt:}
Used when the agent is configured for MCP-only interaction, restricting GUI tools but potentially allowing BASH.
\begin{lstlisting}[caption=API-Only System Prompt, basicstyle=\ttfamily\tiny]
SYSTEM_PROMPT_API_ONLY = f"""<SYSTEM_CAPABILITY>
* You are utilising an Ubuntu virtual machine using {platform.machine()} architecture with internet access.
* You can feel free to install Ubuntu applications with your bash tool. Use curl instead of wget.
* When using your bash tool with commands that are expected to output very large quantities of text, redirect into a tmp file and use str_replace_editor or `grep -n -B <lines before> -A <lines after> <query> <filename>` to confirm output.
* When using your computer function calls, they take a while to run and send back to you.  Where possible/feasible, try to chain multiple of these calls all into one function calls request.
* The current date is {datetime.today().strftime('%A, %B %-d, %Y')}.
</SYSTEM_CAPABILITY>

<IMPORTANT>
* If the item you are looking at is a pdf, if after taking a single screenshot of the pdf it seems that you want to read the entire document instead of trying to continue to read the pdf from your screenshots + navigation, determine the URL, use curl to download the pdf, install and use pdftotext to convert it to a text file, and then read that text file directly with your StrReplaceEditTool.
</IMPORTANT>"""
\end{lstlisting}

\noindent \textbf{3. No-BASH System Prompt:}
Used in configurations where the BASH tool is disabled, but GUI interaction is still allowed.
\begin{lstlisting}[caption=No-BASH System Prompt, basicstyle=\ttfamily\tiny]
SYSTEM_PROMPT_NO_BASH = f"""<SYSTEM_CAPABILITY>
* You are utilising an Ubuntu virtual machine using {platform.machine()} architecture with internet access.
* To open firefox, please just click on the firefox icon.  Note, firefox-esr is what is installed on your system.
* When viewing a page it can be helpful to zoom out so that you can see everything on the page.  Either that, or make sure you scroll down to see everything before deciding something isn't available.
* When using your computer function calls, they take a while to run and send back to you.  Where possible/feasible, try to chain multiple of these calls all into one function calls request.
* The current date is {datetime.today().strftime('%A, %B %-d, %Y')}.
</SYSTEM_CAPABILITY>

<IMPORTANT>
* When using Firefox, if a startup wizard appears, IGNORE IT.  Do not even click "skip this step".  Instead, click on the address bar where it says "Search or enter address", and enter the appropriate search term or URL there.
</IMPORTANT>"""
\end{lstlisting}

\noindent \textbf{4. No-BASH API-Only System Prompt:}
Used in the most restrictive setting where both BASH and direct GUI manipulation tools are disabled, forcing reliance on MCP tools only.
\begin{lstlisting}[caption=No-BASH API-Only System Prompt, basicstyle=\ttfamily\tiny]
SYSTEM_PROMPT_NO_BASH_API_ONLY = f"""<SYSTEM_CAPABILITY>
* You are utilising an Ubuntu virtual machine using {platform.machine()} architecture with internet access.
* When using your computer function calls, they take a while to run and send back to you.  Where possible/feasible, try to chain multiple of these calls all into one function calls request.
* The current date is {datetime.today().strftime('%A, %B %-d, %Y')}.
</SYSTEM_CAPABILITY>
"""
\end{lstlisting}

\paragraph{Other Prompting Components}
In addition to the system prompts, the LLM also receives:
\begin{itemize}
    \item \textbf{Task Instructions:} The specific user instruction for each task (e.g., ``Translate the selected text in this document to French and save it.'') is provided as the initial user message in the conversation history.
    \item \textbf{Tool Descriptions and Schemas:} For each available tool (local or MCP-provided), its name, a natural language description of its functionality, and a JSON schema defining its input parameters are dynamically compiled and presented to the LLM.
\end{itemize}
It is important to note that while these components define the high-level context and capabilities, the detailed turn-by-turn reasoning, the generation of ``thought'' processes within the ReAct cycle, and the specifics of how the LLM translates its intent into a precise tool call (e.g., selecting an MCP tool and filling its parameters based on the schema) are primarily managed by the underlying LLM (Claude 3.7 Sonnet). These lower-level prompting mechanics are not explicitly enumerated as user-defined templates in this work.

\section{Factor Analysis: Impact of BASH Tool Availability}

\label{app:bash_impact_analysis}
\paragraph{Motivation and Context}
A preliminary analysis of our experimental data suggested that the agent frequently and effectively utilized the \texttt{BashTool} to complete various sub-tasks, particularly those involving direct file system operations, application scripting, or information retrieval via command-line utilities. This ease of use for common operations via BASH contrasts with some existing benchmarks where direct BASH access for agents is either limited, not robustly supported, or entirely unavailable, forcing agents in those environments to rely exclusively on GUI or more restricted API interactions for similar functionalities.

Given this observation, and to better understand the extent to which BASH tool availability contributes to the agent's performance in \sys, we conducted a specific analysis focusing on the impact of the \texttt{BashTool}. The goal was to quantify its contribution and to highlight how its presence might differentiate agent capabilities compared to environments with more restricted toolsets.

To isolate the impact of the \texttt{BashTool}, we performed an ablation study. In this study, the \texttt{BashTool} was deliberately removed from the set of tools available to the agent. All other experimental parameters, including the LLM (\texttt{claude-3-7-sonnet-20250219} at temperature 0), the core system prompt, available GUI and MCP tools (but without BASH), and task objectives, remained identical to our primary experimental runs. We then re-evaluated the agent's performance on the full \sys benchmark suite under this BASH-disabled condition. The primary metrics, Task Success Rate (SR) and Key Step Completion Rate (KSCR), were compared against the baseline performance where the \texttt{BashTool} was available.

\paragraph{Results and Observations}
\renewcommand{\arraystretch}{1.5}
\begin{table*}[ht]
\centering
\caption{
Success rates across tasks grouped by difficulty level and modality, under both bash-enabled and bash-disabled conditions.
}
\label{tab:exp_baseline_full}
\small
\scalebox{1}{
\renewcommand{\arraystretch}{1.1} 
\begin{tabularx}{\linewidth}{>{\centering\arraybackslash}p{3cm} >{\centering\arraybackslash}p{3cm} >{\centering\arraybackslash}X >{\centering\arraybackslash}X}
\toprule
\textbf{Difficulty Level} & \textbf{Modality} & \textbf{SR(Bash-enabled)} & \textbf{SR(Bash-disabled)} \\ 
\midrule

Easy & GUI-only & 90.41\% & 91.67\% \\
     & MCP-only & 63.01\% & 55.56\% \\
     & Hybrid   & 90.41\% & 87.50\% \\
\hline

Medium & GUI-only & 72.29\% & 63.86\% \\
       & MCP-only & 51.81\% & 40.96\% \\
       & Hybrid   & 74.70\% & 66.27\% \\
\hline

Hard & GUI-only & 35.56\% & 33.33\% \\
     & MCP-only & 40.00\% & 26.67\% \\
     & Hybrid   & 51.11\% & 28.89\% \\

\bottomrule
\end{tabularx}
}
\end{table*}
\renewcommand{\arraystretch}{1}

\begin{figure}[htbp]
    \centering
    \caption{Comparison of success rate and key step completion rate across different modalities and bash settings.}
    \includegraphics[width=\linewidth]{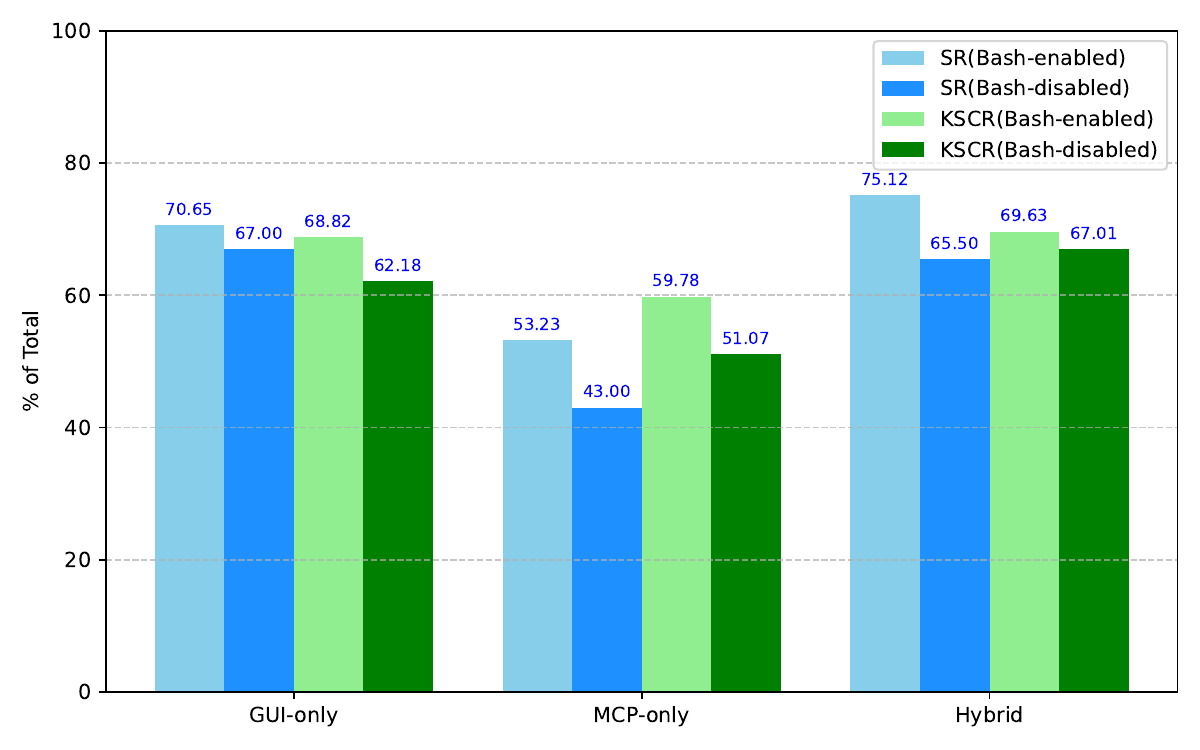}
    \label{fig:sr_comparison}
\end{figure}

The ablation study, removing the \texttt{BashTool}, revealed a clear impact on agent performance across different configurations and task difficulties (see Table~\ref{tab:exp_baseline_full} and Figure~\ref{fig:sr_comparison}).
Overall, enabling the \texttt{BashTool} generally improved SR. For instance, in the Hybrid mode, SR increased from 65.50\% (Bash-disabled) to 75.12\% (Bash-enabled). Similar trends were observed for GUI-Only (67.00\% to 70.65\%) and MCP-Only (43.00\% to 53.23\%) configurations.
The impact varied with task difficulty. For \textbf{Easy} tasks, BASH provided a notable SR boost in MCP-Only mode (63.01\% vs. 55.56\% disabled) and a slight one in Hybrid mode (90.41\% vs. 87.50\% disabled), but slightly decreased SR in GUI-Only mode (90.41\% vs. 91.67\% disabled).
For \textbf{Medium} tasks, BASH consistently improved SR across all modes.
Crucially, for \textbf{Hard} tasks, BASH significantly enhanced SR in MCP-Only (40.00\% vs. 26.67\% disabled) and Hybrid modes (51.11\% vs. 28.89\% disabled), while its impact on GUI-Only mode was marginal (35.56\% vs. 33.33\% disabled).

\paragraph{Discussion}
The general SR improvement with BASH underscores its value in providing agents with a versatile and often more efficient means to interact with the system, particularly for file manipulations, environment setup, or invoking command-line utilities, thus complementing GUI or MCP interactions.
The nuanced results on \textbf{Easy} tasks suggest that for very simple GUI operations, the overhead or complexity of choosing to use BASH might not be beneficial. However, when MCP is involved, even for simple tasks, BASH can fill functional gaps or offer more direct paths.
The substantial gains on \textbf{Hard} tasks, especially in MCP-Only and Hybrid modes, highlight BASH's critical role in tackling complexity. It allows agents to combine MCP's structured application-level control with BASH's system-level flexibility, proving essential where intricate sequences of operations are needed. The limited impact on GUI-Only hard tasks suggests BASH alone may not overcome the inherent challenges of complex GUI automation if robust application-specific APIs (like MCP) are absent.

The divergence between SR and KSCR in the BASH-enabled MCP-Only mode is a direct consequence of an intentional benchmark design choice: \sys allows agents to use tools like BASH to find "shortcuts" to the final goal, potentially bypassing some predefined key step verification points. While this leads to successful task completion (higher SR), it may not follow the detailed sequence of all annotated key steps (lower KSCR).

The findings suggest that when comparing agent performance across different benchmarks, the availability and nature of fundamental tools like shell access should be carefully considered as a potentially significant confounding factor. For \sys, providing BASH access aligns with our goal of enabling agents to leverage a broad spectrum of realistic computer interaction methods. Future research could explore the trade-offs between powerful, general-purpose tools like BASH and a larger set of more specialized, safer, high-level tools that abstract away some of the complexities of direct shell interaction.

\section{Complete Task List}
\label{app:task_list}
\renewcommand{\arraystretch}{1.5}

\begin{longtable}{
>{\RaggedRight\arraybackslash}p{.2in}    
>{\RaggedRight\arraybackslash}p{.6in}      
>{\RaggedRight\arraybackslash}p{2.4in}   
>{\centering\arraybackslash}p{.8in}      
>{\RaggedRight\arraybackslash}p{.6in}     
}

\caption{List of all tasks in \sys.} \label{tab:all-task-list} \\
\toprule
\textbf{ID} & \textbf{App Name} & \textbf{Template} & \textbf{Key Steps} & \textbf{Difficulty Level} \\
\endhead

\hline
1 & VScode & Change the color theme to \{theme\} in VScode. & 1 & Easy \\
\hline
2 & VScode & In the current VScode workspace, find and open the file named \{filename\}, replace \{originname\} with \{expectedname\}, and make sure to save the file. & 2 & Medium \\
\hline
3 & VScode & Open the file named \{filename\} in the current VScode workspace and fold all code blocks. & 2 & Easy \\
\hline
4 & VScode & Modify or create the .vscode/settings.json configuration file in the current VScode workspace, set the font size to \{fontsize\}, and make sure to save the file. & 2 & Medium \\
\hline
5 & VScode & Open the terminal in the current VScode workspace, change to the \{dir\} directory, and execute the command \{cmd\}. & 2 & Medium \\
\hline
6 & VScode & Commit all uncommitted changes in the current VScode workspace with the commit message \{commitmessage\}. & 2 & Easy \\
\hline
7 & VScode & Attach a new window to the current VScode window and open the directory \{workspacedir\} in the new window. & 1 & Medium \\
\hline
8 & VScode & Open the file named \{filename\}, set breakpoints as specified in \{breakpoints\}, and start debugging using the 'c++ debug' configuration. & 6 & Hard \\
\hline
9 & VScode & Search for the definition of the \{operator\} overload in class \{classname\}, and set a breakpoint at the return statement without starting a debug session. & 2 & Medium \\
\hline
10 & VScode & Search for the plugin named \{pluginname\} in VScode and install it. & 1 & Easy \\
\hline
11 & VScode & Set the auto-save mode of the current VScode workspace to \{mode\} and the delay to \{delay\}. & 1
 & Medium \\
\hline
12 & VScode & Switch the workspace to \{directory\}, configure the launch settings for \{filename\}, open the file, set breakpoints as \{breakpoints\}, and continue until \{state\}. & 6 & Hard \\
\hline
13 & VScode & Find and open the file named \{filename\}, place the cursor at the definition of \{functionname\}. & 2 & Easy \\
\hline
14 & VScode & Set the tab size for the current workspace in VScode to \{tabsize\}. & 2 & Medium \\
\hline
15 & VScode & Open \{firstfile\} and \{secondfile\} side-by-side in split view in VScode. & 3 & Medium \\
\hline
16 & VScode & Create a new empty file \{file\} in the current VScode workspace. If the directory does not exist, create it. & 1 & Medium \\
\hline
17 & VScode & In the current VScode workspace, rename all files with suffix \{originsuffix\} under \{relativepath\} to have suffix \{expectedsuffix\}. & 1 & Hard \\
\hline
18 & VScode & Install the plugin \{pluginname\} in VScode and then \{instruction\}. & 2 & Medium \\
\hline
19 & VScode & Use VScode to open \{filename1\} and \{filename2\}, then compare their contents. & 1 & Medium \\
\hline
20 & VScode & Open the file \{file\} in the current workspace and remove all comments while keeping other content unchanged. & 2 & Hard \\
\hline
21 & VScode & Switch the Git branch of the current workspace in VScode to \{branchname\}. If there are uncommitted changes, commit them with message \{commitmessage\}. & 1 & Medium \\
\hline
22 & VScode & Enable Git extension's blame editor decoration feature in VScode. & 1 & Easy \\
\hline
23 & VScode & Open the file named \{filename\}, understand and fix all errors (e.g., name issues, typos, namespace errors) until no errors remain. & 1 & Hard \\
\hline
24 & VScode & Open the file named \{filename\} and convert all variable and function names to PascalCase. & 1 & Hard \\
\hline
25 & VScode & Open the file named \{filename\} and sort all import statements alphabetically. & 1 & Hard \\
\hline
26 & Zotero & In the opened Zotero, move all categories containing tag \{tag\} to RDF file format and export Note to \{exportpath\} directory, use the full paper title as filename & 1 & Medium \\
\hline
27 & Zotero & In Zotero, move all categories containing tag \{tag\} to category \{movecollectiondesc\}, if not exist then create first & 1 & Hard \\
\hline
28 & Zotero & In the opened Zotero, add to item with doi \{itemdesc\} the attachment at path \{pdfpath\} & 1 & Medium \\
\hline
29 & Zotero & In the opened Zotero in MyLibrary under virtualization, add a subcollection \{subcollectionname\}, if virtualization not exist then create first & 1 & Easy \\
\hline
30 & Zotero & In the opened Zotero, create bibliography for entry whose title contains \{itemdesc\}, export to path \{exportpath\}, format as RTF & 1 & Medium \\
\hline
31 & Zotero & Adjust font size of opened Zotero client to be larger & 1 & Medium \\
\hline
32 & Zotero & In Zotero, read PDF content of entry whose title contains \{itemdesc\}, search for GitHub URLs and import to Note & 1 & Hard \\
\hline
33 & Zotero & Add related entry to the entry whose title contains \{targetitemdesc\}, the related one has doi 10.1145/3616871 and title contains Self Adaptive & 1 & Medium \\
\hline
34 & Zotero & Read PDF content of entry whose title contains \{itemdesc\}, extract abstract keywords/Index Terms and add as tags & 1 & Hard \\
\hline
35 & Zotero & In the opened Zotero MyLibrary, move entry in virtualization whose title contains \{itemdesc\} to collection \{collectiondesc\}, if not exist then create it & 1 & Easy \\
\hline
36 & Zotero & In the opened Zotero, add entry by doi \{itemurl\} to group \{collectionname\}, create if not exist & 1 & Easy \\
\hline
37 & Zotero & Read text file \{doisfilepath\} (one doi per line), import all doi URLs to Zotero category \{collectiondesc\}, create if not exist & 1 & Hard \\
\hline
38 & Zotero & In the opened Zotero, adjust PDF reader to \{pdfreaderpath\} & 1 & Medium \\
\hline
39 & Zotero & In the opened Zotero, restore entry whose title contains \{itemdesc\} from Trash & 1 & Easy \\
\hline
40 & Zotero & In the opened Zotero MyLibrary, add category \{collectionname\} & 1 & Easy \\
\hline
41 & Zotero & In the opened Zotero, add tag \{tag\} to entry whose title contains \{itemdesc\} & 1 & Easy \\
\hline
42 & Zulip & In Zulip, under the topic in test Channel, send a scheduled message with content: \{messageinfo\}, set the scheduled time to \{scheduletime\}. & 3 & Medium \\
\hline
43 & Zulip & In Zulip, mute the \{channelname\} Channel. & 1 & Easy \\
\hline
44 & Zulip & In Zulip, pin the \{channelname\} Channel to the top. & 1 & Easy \\
\hline
45 & Zulip & In Zulip, set the user status to \{userstatus\}. & 2 & Easy \\
\hline
46 & Zulip & In Zulip, add an emoji reaction \{emoji\} to your latest message. & 2 & High \\
\hline
47 & Zulip & In Zulip, get the latest message in the \{sourcechannel\} Channel's \{sourcetopic\} topic, and send this message to the \{targetchannel\} Channel's \{targettopic\} topic. & 2 & Medium \\
\hline
48 & Zulip & In the \{channelname\} Channel of Zulip, create a new topic named \{topicname\} with content \{topicmessage\}. & 3 & Medium \\
\hline
49 & Zulip & In Zulip, under the general Channel's \{targettopic\} topic (create it if it doesn't exist), add an emoji reaction \{targetemoji\} to the latest message. & 5 & High \\
\hline
50 & Zulip & In Zulip, list all topics in \{sourcechannel\} Channel, join them into one string with space separator, and send it to \{targetchannel\} Channel's \{targettopic\} topic. & 3 & Medium \\
\hline
51 & Zulip & In Zulip, unsubscribe from the \{channelname\} Channel. & 1 & Easy \\
\hline
52 & Zulip & In Zulip, send an organization invitation to the user with email \{email\}. The user joins with role \{role\}, and the invitation never expires. & 3 & Medium \\
\hline
53 & Zulip & In Zulip, check if you are subscribed to the \{channelname\} Channel. If not, subscribe to it. & 2 & Easy \\
\hline
54 & Zulip & In Zulip, create a new Channel with name \{channelname\} and description \{description\}, subscribe all users by default, leave other settings default. & 3 & Medium \\
\hline
55 & Zulip & In Zulip, send a direct message to \{email\} with content: 'have a nice day'. & 2 & Easy \\
\hline
56 & Zulip & In Zulip, send a direct message to \{username\} with content \{message\}. & 2 & Easy \\
\hline
57 & Zulip & In Zulip, set user online status to \{onlinestatus\}. & 2 & Easy \\
\hline
58 & Zulip & In Zulip, subscribe to the messages in the \{channelname\} Channel. & 2 & Easy \\
\hline
59 & Zulip & In Zulip, send message \{message\} to topic \{topicname\} in channel \{channelname\}. & 2 & Medium \\
\hline
60 & QGIS & Adjust the layer order in QGIS, the order is \{LayerOrder\} (top-down, separated by commas). & 1 & Hard \\
\hline
61 & QGIS & Set the project CRS to \{crs\} in QGIS. & 1 & Medium \\
\hline
62 & QGIS & Set the color of the layer \{LayerName\} to \{color\} (RGB hex) in QGIS. & 2 & Hard \\
\hline
63 & QGIS & Set the unit of the layer \{LayerName\} to \{unit\} (Options: Millimeters, Points, Pixels, Meters at Scale, Map Units, Inches) in QGIS. & 2 & Medium \\
\hline
64 & QGIS & Export the map as an image in QGIS, width: \{width\} px, height: \{height\} px, save path: \{savepath\}, format: \{type\}. & 2 & Hard \\
\hline
65 & QGIS & Set the opacity of the layer \{LayerName\} to \{opacity\}\% (0-100\%) in QGIS. & 2 & Medium \\
\hline
66 & QGIS & Remove a layer in QGIS, layer name: \{LayerName\}. & 1 & Easy \\
\hline
67 & QGIS & Add a layer note to the layer \{LayerName\} in QGIS, content: \{notes\}. & 2 & Easy \\
\hline
68 & QGIS & Set the CRS of the layer \{layername\} to \{crs\} in QGIS. & 2 & Medium \\
\hline
69 & QGIS & Add a raster layer in QGIS, absolute path: \{rasterLayerPath\}. & 1 & Easy \\
\hline
70 & QGIS & Add a vector layer in QGIS, absolute path: \{vectorLayerPath\}. & 1 & Easy \\
\hline
71 & QGIS & Zoom the map canvas to the extent of the specified layer in QGIS, layer name: \{LayerName\}. & 1 & Easy \\
\hline
72 & QGIS & Export the map as a PDF in QGIS, width: \{width\} px, height: \{height\} px, save path: \{savepath\}. & 2 & Hard \\
\hline
73 & QGIS & Create and save a new project in QGIS, path: \{savepath\}. & 2 & Hard \\
\hline
74 & QGIS & Load a project in QGIS, path: \{loadpath\}. & 1 & Easy \\
\hline
75 & obs-studio & Please add a new scene in obs-studio named \{newscenename\} & 1 & Easy \\
\hline
76 & obs-studio & Please start recording in obs-studio & 1 & Easy \\
\hline
77 & obs-studio & Please set the streaming service to \{streamservice\} in obs-studio & 1 & Medium \\
\hline
78 & obs-studio & Please rename \{oldscenename\} to \{newscenename\} in obs-studio & 1 & Easy \\
\hline
79 & obs-studio & Please switch to scene \{scene\} in obs-studio & 1 & Easy \\
\hline
80 & obs-studio & Please add a new text source in the scene \{scenename\} in obs, named \{newsourcename\}, with settings \{settings\} & 3 & Hard \\
\hline
81 & obs-studio & Please stop recording in obs-studio & 1 & Easy \\
\hline
82 & obs-studio & Set OBS transition animation to \{transitionname\}, set duration to \{durationms\} milliseconds, and switch scene to \{destscene\} & 2 & Hard \\
\hline
83 & obs-studio & Please delete a scene named \{delscenename\} in obs-studio & 1 & Easy \\
\hline
84 & obs-studio & Please change the transition animation to \{transition\} in obs-studio & 1 & Easy \\
\hline
85 & obs-studio & Current DISPLAY is defined in \~{}/.bashrc, Do not launch another obs-studio if one is already launched. Add a Stinger transition effect in OBS Studio, using \{stingerfile\} as the video file, set the transition point to \{transitionpointms\} milliseconds, then demonstrate switching between two scenes & 3 & Hard \\
\hline
86 & obs-studio & Please export the profile named \{profilename\} in obs to \{exportpath\}, then import the profile configuration file from \{importpath\} & 2 & Medium \\
\hline
87 & obs-studio & Please create a new scene in obs-studio named \{newscenename\}, and add a new source to this scene named \{sourcename\}, with type \{sourcetype\}. Set color source width to \{width\} and height to \{height\} & 4 & Medium \\
\hline
88 & obs-studio & Please batch add color sources \{addcolorsources\} in \{scenename\} in obs-studio, then reorder them from bottom to top as \{reorderto\}, and finally delete \{deletesources\} & 3 & Hard \\
\hline
89 & obs-studio & Please add \{filters\} filters to the text source named '\{textsourcename\}' in obs-studio, then disable, enable and remove them & 3 & Medium \\
\hline
90 & obs-studio & Please add a new display capture source in obs-studio named \{newsourcename\}, with type \{newsourceid\} & 1 & Medium \\
\hline
91 & obs-studio & Please export a screenshot of scene \{scenename\} in obs-studio, save it to \{savepath\}, with resolution \{width\}x\{height\} & 2 & Medium \\
\hline
92 & obs-studio & Please set the shortcut key for starting and ending recording in OBS to \{hotkey\}, and test the shortcut key by triggering it & 3 & Medium \\
\hline
93 & obs-studio & Please create a new scene \{newscenename\} in obs-studio, and add a new looped video source in the new scene, with name \{sourcename\} and type \{sourcetype\} & 4 & Medium \\
\hline
94 & obs-studio & Please create a new scene collection in obs-studio named \{collectionname\}, and create scenes \{scenenames\} inside it & 2 & Medium \\
\hline
95 & obs-studio & Add an image source named '\{sourcename\}' in obs-studio, using the image '\{imagepath\}', and set its opacity to \{opacity\}\% & 2 & Hard \\
\hline
96 & obs-studio & Delete multiple or single color sources named \{sourcenames\} in obs-studio & 0 & Medium \\
\hline
97 & obs-studio & Use a profile in obs-studio, set the video bitrate to \{videobitrate\}Kbps, audio bitrate to \{audiobitrate\}, encoder preset to \{encoderpreset\} in the output streaming settings with output mode Simple, and enable Replay Buffer with maximum replay time of \{replaybuffertime\} seconds & 5 & Hard \\
\hline
98 & obs-studio & Please set the visibility of \{sourcename\} to \{sourcevisible\} in obs-studio & 1 & Easy \\
\hline
99 & obs-studio & Please configure the recording output path as \{outputpath\} in obs-studio, with format \{outputformat\}, and then test the recording function & 3 & Medium \\
\hline
100 & Joplin & Change the note title from \{oldnotename\} to \{newnotename\} under notebook \{notebookname\} in Joplin & 1 & Medium\\
\hline
101 & Joplin & Add the following tags \{tagnames\} to the note named \{notename\} in Joplin & 1 & Medium\\
\hline
102 & Joplin & Set the due date of the todo named \{todoname\} in Joplin to \{duedate\} & 1 & Medium\\
\hline
103 & Joplin & Change the sorting of the current note page to descending by title in Joplin & 1 & Hard\\
\hline
104 & Joplin & Add the tag \{tagname\} to the following notes \{notenames\} in Joplin & 1 & Medium\\
\hline
105 & Joplin & Add the PDF file \{pdffilepath\} to the note \{notetitle\} under notebook \{notebookname\} in Joplin & 1 & Hard\\
\hline
106 & Joplin & Restore the note named \{notename\} from trash in Joplin & 1 & Medium\\
\hline
107 & Joplin & Move the note named \{notename\} to notebook \{targetnotebook\} in Joplin & 1 & Easy\\
\hline
108 & Joplin & Append the following text to the note named \{notename\} in Joplin: \{appendtext\} & 1 & Medium\\
\hline
109 & Joplin & Set the zoom level of the view to \{zoomlevel\} \% in Joplin & 1 & Medium\\
\hline
110 & Joplin & Create a notebook named \{notebookname\} in Joplin (not a note) & 1 & Easy\\
\hline
111 & Joplin & Tag the note named \{notename\} with \{tagname\} in Joplin & 1 & Easy\\
\hline
112 & Joplin & Delete the todo item titled \{todocontent\} in Joplin & 1 & Medium\\
\hline
113 & Joplin & Change the color theme to \{theme\} in Joplin & 1 & Easy\\
\hline
114 & Joplin & Create a todo titled \{todocontent\} in Joplin & 1 & Easy\\
\hline
115 & Joplin & Delete the note named \{notename\} in Joplin & 1 & Easy\\
\hline
116 & Joplin & Create a note named \{notename\} in the notebook \{notebookname\} in Joplin & 1 & Easy\\
\hline
117 & Joplin & Add location info to the note named \{notename\} in Joplin: latitude \{latitude\}, longitude \{longitude\} & 1 & Hard\\
\hline
118 & Joplin & Open the note named \{notename\} in the current window in Joplin & 1 & Easy\\
\hline
119 & Joplin & Mark the todo titled \{todocontent\} as completed via checkbox in Joplin & 1 & Easy\\
\hline
120 & Joplin & Change the current interface to rich text editor mode in Joplin & 1 & Easy\\
\hline
121 & Joplin & Import the file \{markdownfilepath\} into the notebook \{notebookname\} in Joplin & 1 & Hard\\
\hline
122 & FreeCAD & The application is already open. In FreeCAD, create an additive cylinder with radius \{radius\} and height \{height\}. Create it using 'create an additive cylinder', and save the file to \{sourcepath\}\{filename\} & 2 & Medium\\
\hline
123 & FreeCAD & The application is already open. Open a document in FreeCAD application, the document name is \{sourcepath\}\{filename\}, show the opening result & 1 & Easy\\
\hline
124 & FreeCAD & The application is already open. Create and save a document in FreeCAD application, the document name should be \{sourcepath\}\{filename\}, show the save result & 1 & Easy\\
\hline
125 & FreeCAD & The application is already open. In FreeCAD, create an additive sphere with radius \{radius\} (typeid = additive sphere). Create it using 'create an additive sphere', and save the file to \{sourcepath\}\{filename\} & 2 & Medium\\
\hline
126 & FreeCAD & The application is already open. In the FreeCAD application, first set the user edit mode to \{firstmode\}, then set it back to \{finalmode\} & 2 & Medium\\
\hline
127 & FreeCAD & The application is already open. Create a new document in FreeCAD application and show the result & 1 & Easy\\
\hline
128 & FreeCAD & The application is already open. In FreeCAD, add an additive cube with length \{length\}, width \{width\}, height \{height\}. Create it using 'create an additive Box', and save the file to \{sourcepath\}\{filename\} & 2 & Medium\\
\hline
129 & FreeCAD & The application is already open. In FreeCAD's part design, create a line with length \{linelength\} through a sketch, and save the file to \{sourcepath\}\{filename\} & 2 & Medium\\
\hline
130 & FreeCAD & The application is already open. Create a cylinder with radius \{cylinderradius\}mm and height \{cylinderheight\}mm in FreeCAD, then create a tilted rectangular groove on its side. The groove should have width \{groovewidth\}mm, depth \{groovedepth\}mm, height \{grooveheight\}mm, and be tilted around the Y-axis by \{grooveangle\} degrees. The cylinder should be created as an additive cylinder (typeid), and the rectangular groove should be created as a subtractive box (typeid). Do not create the shapes separately and use Boolean subtraction or cut. Finally, save the file to \{sourcepath\}\{filename\} & 2 & Hard\\
\hline
131 & FreeCAD & The application is already open. Create a regular polygon in FreeCAD's part design using sketch with \{sides\} sides and a circumscribed circle radius of \{radius\}mm, then save the file to \{sourcepath\}\{filename\} & 2 & Hard\\
\hline
132 & FreeCAD & The application is already open. In the FreeCAD application, using open first start setup, first set the unit system to no.\{firstunitsystem\} which called \{finalunitsystemname\}, then set it back to no.\{finalunitsystem\} which called \{finalunitsystemname\} & 2 & Medium\\
\hline
133 & FreeCAD & The application is already open. Create a cuboid in FreeCAD with length \{cubelength\}, width \{cubewidth\}, and height \{cubeheight\}, then create a wedge-shaped cut on its top with parameters: Xmin=\{wedgeXmin\}, Xmax=\{wedgeXmax\}, Ymin=\{wedgeYmin\}, Ymax=\{wedgeYmax\}, Zmin=\{wedgeZmin\}, Zmax=\{wedgeZmax\}, X2min=\{wedgeX2min\}, X2max=\{wedgeX2max\}, Z2min=\{wedgeZ2min\}, Z2max=\{wedgeZ2max\}. The cuboid should be created using 'create an additive cube', and the wedge cut should be created using 'create subtractive wedge'. Finally, save the file to \{sourcepath\}\{filename\} & 2 & Hard\\
\hline
134 & FreeCAD & The application is already open. In FreeCAD, create a triangular prism with an equilateral triangle as its base, with a circumradius of \{prismcircumradius\}mm and a height of \{prismheight\}mm. Then add a stepped hole in the center of its top face. The stepped hole consists of two parts: an inner small hole with a radius of \{holeinnerradius\}mm that goes through the entire prism, and an outer larger hole with a radius of \{holeouterradius\}mm that extends from the top face to a depth of \{holedepth\}mm. The triangular prism can be created by 'create an additive prism' (typeid), and the stepped hole should be created using subtractive cylinders (typeid). Do not create the shapes separately and use boolean subtraction or cut. Finally, save the file to \{sourcepath\}\{filename\} & 2 & Hard\\
\hline
135 & FreeCAD & The application is already open. In FreeCAD's 2D draft, create a square with center at origin (0,0) and side length of \{sidelength\}, then save the file to \{sourcepath\}\{filename\} & 2 & Medium\\
\hline
136 & FreeCAD & The application is already open. In the FreeCAD application, using open first start setup, first change the interface language to \{firstlanguage\}, then change it back to \{finallanguage\} & 2 & Medium\\
\hline
137 & FreeCAD & The application is already open. In FreeCAD, create a cone with a bottom radius of \{coneradius1\}mm, a top radius of \{coneradius2\}mm, and a height of \{coneheight\}mm. Then, create a hemispherical cutout at the center of its top surface. The radius of the cutting sphere is \{sphereradius\}mm, and its center should be positioned at the cone's top surface center (\{spherepositionx\}, \{spherepositiony\}, \{spherepositionz\})mm. The cone should be created using the 'create an additive cone' method (typeid), and the hemispherical cutout should be created using the 'create subtractive sphere' method (typeid). Do not create the shapes separately and use Boolean subtraction or cut. Finally, save the file to \{sourcepath\}\{filename\}. & 2 & Hard\\
\hline
138 & FreeCAD & The application is already open. Create a pyramid in FreeCAD with a rectangular base that has length \{baselength\}mm and width \{basewidth\}mm, and a height of \{pyramidheight\}mm. The base should be positioned on the XZ plane. Then add a mirror feature to this pyramid using the \{mirrorplane\} plane. The pyramid can be created using 'create a addtive wedge'. The mirror feature should be created using 'apply a pattern -> mirror'. Finally, save the file to \{sourcepath\}\{filename\} & 2 & Hard\\
\hline
139 & FreeCAD & The application is already open. In FreeCAD's 2D draft, create a circle with center at origin (0,0) and radius of \{radius\}, then save the file to \{sourcepath\}\{filename\} & 2 & Medium\\
\hline
140 & FreeCAD & The application is already open. Create an octagonal prism in FreeCAD with a circumscribed circle radius of \{prismradius\}mm and height of \{prismheight\}mm, then add a circular through-hole with radius \{holeradius\}mm in the center of the prism. The prism can be created using 'create a additive prism'(typeid), and the circular hole should be created using 'create a subtractive cylinder'(typeid). Do not create the shapes separately and use Boolean subtraction or cut. Finally, save the file to \{sourcepath\}\{filename\} & 2 & Hard\\
\hline
141 & FreeCAD & The application is already open. Create a disk in FreeCAD with radius \{diskradius\}mm and height \{diskheight\}mm, then add a threaded hole in its center with the following parameters:\newline
- Profile: ISO metric regular profile\newline
- Size: \{threadsize\}\newline
- Clearance: Standard\newline
- Depth: \{threaddepth\}mm\newline
- Threaded: checked\newline
- Model Thread: \{modelthread ? 'checked' : 'unchecked'\}\newline
The disk should be created as an additive cylinder (typeid), and the threaded hole should be created using the create a threaded hole (typeid) method. Do not create the shapes separately and use boolean subtraction or cut. Finally, save the file to \{sourcepath\}\{filename\} & 2 & Hard\\
\hline
142 & FreeCAD & The application is already open. Create a cube in FreeCAD with length \{cubelength\}mm, width \{cubewidth\}mm, and height \{cubeheight\}mm, then add fillets with a radius of \{filletradius\}mm to all its edges. The cube should be created as an additive cube (typeid), and the fillets should be created using the create a fillet feature on cube. Do not create the shapes separately and use Boolean subtraction or cut. Finally, save the file to \{sourcepath\}\{filename\} & 2 & Hard\\
\hline
143 & FreeCAD & The application is already open. In FreeCAD's Part Design workbench, create a rounded rectangle with length \{length\} mm, width \{width\} mm, and corner radius \{radius\} mm in a sketch, and save the file to \{sourcepath\}\{filename\} & 2 & Medium\\
\hline
144 & FreeCAD & The application is already open. Create a disk in FreeCAD with radius \{diskradius\}mm and height \{diskheight\}mm, then create a toroidal groove on its top surface. The toroidal groove's major radius (Radius1) is \{torusradius1\}mm, minor radius (Radius2) is \{torusradius2\}mm, with angle parameters Angle1=\{torusangle1\} degrees, Angle2=\{torusangle2\} degrees, and Angle3=\{torusangle3\} degrees, defaulting to a complete toroidal groove. The groove depth on the disk equals the minor radius. The disk should be created using 'create an additive cylinder' (typeid), and the toroidal groove should be created using 'create subtractive torus' (typeid). Do not create the shapes separately and use boolean subtraction. Finally, save the file to \{sourcepath\}\{filename\} & 2 & Hard\\
\hline
145 & FreeCAD & The application is already open. Create a cylinder with radius \{cylinderradius\} and height \{cylinderheight\}, and create \{holecount\} holes with radius \{holeradius\} evenly distributed on its side. These holes should be evenly spaced along the height of the cylinder, and the center of each hole should maintain the same horizontal distance from the cylinder's axis. The cylinder should be created using 'create an additive cylinder', and the holes should be created using 'create subtractive cylinder'. Finally, save the file to \{sourcepath\}\{filename\} & 2 & Hard\\
\hline
146 & Anki & Add a Q\&A card to the default system deck, front side is "To be or not to be" with "be" and "not to be" colored red, back side is "that's a question", add tag "quote" & 3 & Hard \\
\hline
147 & Anki & Delete all cards with tag \{tagname\} & 2 & Medium \\
\hline
148 & Anki & Create a card with front content "explain quick sort algorithm", with tag \{tagname\} & 2 & Medium \\
\hline
149 & Anki & Add a Q\&A card with question "Three largest countries in the world", bold "largest", answer as an ordered list with Russia, Canada, China & 2 & Medium \\
\hline
150 & Anki & Add a deck named \{deckname\}, delete it, then undo the deletion & 2 & Easy \\
\hline
151 & Anki & Add a cloze deletion card to the default deck, content "The capital of China is Beijing", with Beijing clozed out & 2 & Medium \\
\hline
152 & Anki & Add a Q\&A card to the default deck, front side "x\^2-2x+1=0" using superscript, and mark "x\^2" font color red, back side "x=1" & 2 & Medium \\
\hline
153 & Anki & Create a deck named \{tagname\}, move all cards with tag \{tagname\} into it & 3 & Hard \\
\hline
154 & Anki & Create a card with front "What is a \{searchkeyword\}", find all cards containing \{searchkeyword\}, set their due date to \{duedays\} days & 3 & Medium \\
\hline
155 & Anki & Add red flags to all cards with tag \{tagname\} & 2 & Medium \\
\hline
156 & Anki & Set all cards with tag \{tagname\} to suspended & 1 & Easy \\
\hline
157 & Anki & Create a deck named \{deckname\} and then delete it & 2 & Easy \\
\hline
158 & Anki & Set the default search box content to "cs" & 1 & Easy \\
\hline
159 & Anki & Create a template named \{name\} by copying the Q\&A template. Add a new field called Note, place it in the third field position. Use conditional replacement syntax on the front side, for example: \{Tags\} Tag: \{Tags\} \{Tags\}. Using conditional syntax, if the Note field exists, highlight the front side and put Note after the answer on the back side separated by <br>, displayed as Note:\{Note\}. If no Note, do not display this line. Add a card with front \{firstfield\}, back \{secondfield\}, Note \{thridfield\} & 4 & Hard \\
\hline
160 & Anki & Add a Q\&A card with front \{firstfield\} and back \{secondfield\}, add it to deck \{deckname\} (create if missing), then undo add & 2 & Medium \\
\hline
161 & Anki & Add a Q\&A card with front \{firstfield\} and back \{secondfield\}, add to default deck & 2 & Easy \\
\hline
162 & Anki & Create a template named \{name\} with 3 fields: question, short answer, long answer. The second and third fields appear on back separated by two <br>. The second field uses red color & 2 & Hard \\
\hline
163 & Anki & Delete all unused tags & 1 & Easy \\
\hline
164 & Anki & Create a deck named \{deckname\} & 2 & Easy \\
\hline
165 & Anki & Use enhanced cloze plugin template (https://ankiweb.net/shared/info/1990296174). Configure plugin: set show Hints For Pseudo Clozes to False. Add a card using this template, front "The quick brown fox jumps over the lazy dog.", cloze brown (c1), fox (c2), dog (c3), cloze lazy always shown (marked). Explanation about display mechanism follows. & 2 & Hard \\
\hline
166 & Anki & In settings, set the maximum advance study time to \{time\} minutes & 1 & Easy \\
\hline
167 & Anki & Set the day start time (daily card rotation time) to \{time\} o'clock & 1 & Easy \\
\hline
168 & Anki & Add a card with front side "what is this?" with a dog image, image path is \{path\}, back side is dog & 3 & Medium \\
\hline
169 & Anki & Set auto backup time to \{time\} minutes & 1 & Easy \\
\hline
170 & Anki & Add a card with front side What is this?. Then find all cards containing What and replace it with Where. Use find and batch replace since there could be many entries & 3 & Medium \\
\hline

171 & qBittorrent & In qBittorrent, pause the download of the file \{filename\} in the session list. & 1 & Easy \\
\hline

172 & qBittorrent & In qBittorrent, remove the file \{filename\} from the session list to cancel the download. & 2 & Easy \\
\hline

173 & qBittorrent & In qBittorrent, set the upload speed limit for the torrent \{torrentname\} to \{uploadlimit\} KB/s. & 2 & Medium \\
\hline

174 & qBittorrent & In qBittorrent, add the file \{filename\} from the desktop to the qBittorrent download list. & 2 & Easy \\
\hline

175 & qBittorrent & In qBittorrent, set the download speed limit for the torrent \{torrentname\} to \{downloadlimit\} KB/s only. & 2 & Medium \\
\hline

176 & qBittorrent & In qBittorrent, set the global download speed limit to \{downloadrate\} KB/s. & 3 & Easy \\
\hline

177 & qBittorrent & In qBittorrent, set the global upload speed limit to \{uploadrate\} KB/s. & 3 & Easy \\

\hline
178 & qBittorrent & In qBittorrent, add the tag \{tagname\} to the torrent \{torrentname\}. & 2 & Medium \\
\hline

179 & Blender & In Blender, open the specified file \{filedir\}/\{filename\}. & 1 & Easy \\
\hline
180 & Blender & In Blender, create a new window. & 1 & Easy \\
\hline
181 & Blender & In Blender, save the file as \{filedir\}/\{filename\}. & 1 & Easy \\
\hline
182 & Blender & In Blender, import the OBJ file \{filedir\}/\{filename\}. & 1 & Easy \\
\hline
183 & Blender & In Blender, add a \{type\} type \{mltype\}. & 1 & Easy \\
\hline
184 & Blender & In Blender, delete existing elements. & 1 & Easy \\
\hline
185 & Blender & In Blender, scale the current Cube element xyz to \{scale\}. & 1 & Easy \\
\hline
186 & Blender & In Blender, add a \{type\} type \{meshtype\} and change view mode to Edit Mode. & 4 & Easy \\
\hline
187 & Blender & In Blender, add a \{type\} type \{meshtype\}, then switch to top view, add text, switch to edit view, and change the text to \{text\}. & 6 & Medium \\
\hline
188 & Blender & In Blender, add a \{type\} type \{meshtype\}, set its origin to Center of Mass (Surface). & 3 & Medium \\
\hline
189 & Blender & In Blender, add a \{type\} type \{meshtype\} and scale its xyz to \{scale\}. & 3 & Medium \\
\hline
190 & Blender & In Blender, clear the location of the existing Metaball type Cube element. & 0 & Medium \\
\hline
191 & Blender & In Blender, change the name of an existing Metaball type Cube element to \{name\} and its size to \{X\}*\{Y\}*\{Z\}. & 3 & Medium \\
\hline
192 & Blender & In Blender, delete an existing Metaball type Cube element, add a \{type\} type \{meshtype\}, and scale its xyz to \{scale\}. & 4 & Medium \\
\hline
193 & Blender & In Blender, add a \{type\} type \{meshtype1\} and \{meshtype2\}, and join them. & 6 & Medium \\
\hline
194 & Blender & In Blender, add a \{type\} type \{meshtype1\} and \{meshtype2\}, set \{meshtype1\} above \{meshtype2\} with Location \{location\}, then join them. & 7 & Hard \\
\hline
195 & Blender & In Blender, add a \{type\} type \{meshtype1\} and \{meshtype2\}, and switch one to local view. & 5 & Easy \\
\hline
196 & Blender & In Blender, add a \{type\} type \{meshtype\}, add a material to it, and switch view type to MATERIAL. & 4 & Medium \\
\hline
197 & Blender & In Blender, add a \{type\} type \{meshtype\}, switch to local view, add material, and change view type to MATERIAL. & 5 & Hard \\
\hline
198 & Blender & In Blender, add a \{type\} type \{meshtype\}, edit vertex (1, 1, 0) to (2, 2, 0) and (-1, -1, 0) to (-2, -2, 0). & 6 & Hard \\
\hline
199 & Blender & In Blender, add a \{type\} type \{meshtype\}, apply a \{modifiertype\} Modifier. & 4 & Medium \\
\hline
200 & Blender & In Blender, add a \{type\} type \{meshtype\}, move it down 2m along Z axis, apply a \{modifiertype\} Modifier, and save as \{filedir\}/\{filename\}. & 7 & Hard \\
\hline
201 & Blender & In Blender, open the file \{filedir\}/\{filename\} and import all OBJ files in the folder. & 4 & Medium \\

\bottomrule
\label{tab:all-task-list}
\end{longtable}

\renewcommand{\arraystretch}{1}



\end{document}